\documentclass[10pt,journal,compsoc]{IEEEtran}
\ifCLASSOPTIONcompsoc
\usepackage{cite}
  \usepackage{times}
\usepackage{epsfig}
\usepackage{graphicx}
\usepackage{amsmath}
\usepackage{amssymb}
\usepackage{booktabs, multicol, multirow}
\usepackage{array}
\usepackage{xcolor}
\usepackage[colorlinks=true]{hyperref}
\usepackage[caption=false,font=footnotesize]{subfig}
\usepackage{float}
\usepackage{epstopdf}
\usepackage{mathtools}
\usepackage{setspace}
\usepackage{balance}
\usepackage{soul}
\usepackage[export]{adjustbox}

\else
  \usepackage{cite}
\fi

\ifCLASSINFOpdf
\else
\fi
\begin{document}
%
\title{Deep GAN-Based Cross-Spectral Cross-Resolution Iris Recognition}
%
%
%
%

\author{Moktari Mostofa,~\IEEEmembership{Student Member,~IEEE,}
        Salman Mohamadi,~\IEEEmembership{}\\ Jeremy Dawson,~\IEEEmembership{member,~IEEE,}
        and~ Nasser M. Nasrabadi,~\IEEEmembership{Fellow,~IEEE}}
\markboth{IEEE TRANSACTIONS ON BIOMETRICS, BEHAVIOR, AND IDENTITY SCIENCE}%
{Shell \MakeLowercase{\textit{et al.}}: Bare Demo of IEEEtran.cls for IEEE Journals}
%



\IEEEtitleabstractindextext{%
\begin{abstract}
In recent years, cross-spectral iris recognition has emerged as a promising biometric approach to establish the identity of individuals. However, matching iris images acquired at different spectral bands (i.e., matching a visible (VIS) iris probe to a gallery of near-infrared (NIR) iris images or vice versa) shows a significant performance degradation when compared to intraband NIR matching. Hence, in this paper, we have investigated a range of deep convolutional generative adversarial network (DCGAN) architectures to further improve the accuracy of cross-spectral iris recognition methods. Moreover, unlike the existing works in the literature, we introduce a resolution difference into the classical cross-spectral matching problem domain. We have developed two different novel techniques using the conditional generative adversarial network (cGAN) as a backbone architecture for cross-spectral iris matching. In the first approach, we simultaneously address the cross-resolution and cross-spectral matching problem by training a cGAN that jointly translates cross-resolution as well as cross-spectral tasks to the same resolution and within the same spectrum. In the second approach, we design a coupled generative adversarial network (cpGAN) architecture consisting of a pair of cGAN modules that project the VIS and NIR iris images into a low-dimensional embedding domain to ensure maximum pairwise similarity between the feature vectors from the two iris modalities of the same subject. To assure the efficacy of our methods, we perform several experiments considering multiple real-life scenarios on three publicly-available cross-spectral iris datasets. Our best experimental results obtained from the cpGAN network outperform the existing benchmark convolutional neural network (CNN) with a supervised discrete hashing (SDH) approach\textcolor{blue}{\cite{wang2019cross}} by as much as 1.67\%, and 2.22\% GAR at FAR of 0.01, while our cGAN provides recognition accuracy with significantly lower EER value of 1.5\%, and 1.54\% for PolyU bi-spectral dataset, and Cross-eyed-cross-spectral iris recognition database, respectively.
It indicates the superiority of our approaches over results previously published in the literature. 
\end{abstract}

\begin{IEEEkeywords}
VIS, NIR, DCGAN, cGAN, cpGAN, CNN.
\end{IEEEkeywords}}

\maketitle

\IEEEdisplaynontitleabstractindextext

%
\IEEEpeerreviewmaketitle

\IEEEraisesectionheading{\section{Introduction}\label{sec:introduction}}

%
%
%
%
 
Identity verification based on the analysis of a person’s physiological properties is believed to be more reliable than other means of identification such as a PIN or password, username, or access card. Fingerprint, palmprint, face, and iris  biometric modalities have seen widespread use for human  identification\textcolor{blue}{\cite{uliyan2020anti,zhao2020deep,dabouei2020boosting,daugman2009iris}}. Among all biological traits, iris is well suited for the most accurate and secure personal authentication because of the distinctive patterns present in the iris textures for individuals\textcolor{blue}{\cite{daugman2009iris, wildes1997iris}}. The human iris pattern is observed to have unique and different textures due to the process of chaotic morphogenesis that causes its formation in early childhood, exhibiting variation even among identical twins. Therefore, in recent decades, iris recognition has received significant attention as an identity verification method in the biometric community\textcolor{blue}{\cite{bowyer2008image, jain201650}} for civilian and surveillance applications.

Conventional iris recognition biometric systems are based on iris images obtained under near-infrared (NIR) illumination due to the optical properties of the human iris in the NIR band of the electromagnetic spectrum. Broadly speaking, the NIR light absorption characteristics of the pigment melanin within the iris tissue determines the visibility of iris texture details in NIR imaging. As a result, the iris textures appear much better under illumination in the $700\sim900$ nm wavelength range compared to illumination within the visible wavelengths in the  $400\sim700$ nm range. For this reason, in most commercial iris recognition systems, single-band near-infrared (NIR) iris matching techniques have been extensively used for identity authentication tasks\textcolor{blue}{\cite{daugman1993high, bowyer2008image}}. These systems use well-established algorithms and protocols to perform identification when the probe and gallery are in the same domain, which has resulted in highly-accurate performance. However, the majority of these methods require close-distance iris imaging to ensure that the acquired images are in good quality with minimum acceptable iris diameter\textcolor{blue}{\cite{daugman2009iris}}. To eliminate these constraints in the NIR-based iris recognition, several visible wavelength based iris recognition systems have been developed\textcolor{blue}{\cite{tan2013towards, zhao2015accurate}} in the last few years, which has expanded the scope of investigating the capabilities of the iris matching techniques under visible light illumination. In addition, several competitions such as the Noisy Iris Challenge Evaluation (NICE)\textcolor{blue}{\cite{bowyer2012results}}, and the Mobile Iris Challenge Evaluation\textcolor{blue}{\cite{de2015mobile}} focus on the realistic acquisition process of visible iris images. The major factors behind this attention to visible wavelength-based iris recognition are (1) visible range cameras are capable of acquiring images from long distance, and (2) they are low-cost compared to NIR cameras. Emerging dual imaging technology in recent smartphones offer image capture in the visible and NIR illuminations. As a result, now police and every law enforcement officer, customs and border protection officer, and special operator has an agency-issued cellphone to perform multi-modal biometric captures (face, fingerprint, and iris), which are used later for authentication. In this context, effective usage of this opportunistic visible iris images requires accurate iris matching with the corresponding NIR images enrolled in the national ID databases.

Moreover, recent advances in video surveillance technology have enabled the capture of very high-resolution iris images in the visible spectrum using low-cost camera technologies, which can be used for identification purposes within the same domain or across different spectra. However, most large-scale galleries of iris images have been acquired in the lower resolution near-infrared (NIR) domain. Therefore, cross-spectral and cross-resolution iris matching has emerged as a major challenge \textcolor{blue}{\cite{wang2019cross, nalla2016toward}}. It is essential to address both cross-spectral as well as cross-resolution methods for matching these opportunistic visible iris query probes against the enrolled NIR iris images in a gallery.

\begin{table*}[t]
\begin{center}
\caption{A summary of the recent related works on cross-spectral iris recognition, accuracy is reported at a given False Acceptance Rate
(FAR).}
\label{tab:my_label}
\scalebox{0.81}{\begin{tabular}{c c c c c c c c}
        \hline
         Reference & Method & Database &  Iris comparison & Iris matching & Features & Matching accuracy \\
        \hline
        \textcolor{blue}{\cite{zuo2010cross}} & A predictive NIR iris
image  & WVU Multi-spectral iris 
& Genuine = 280
& Cross-spectral &Hand-crafted &95.2\%
\\
&is used from the color image & \hspace{-2cm}database & \hspace{0.6cm}Impostor = 20,745 & & & (FAR = 0.001)\\
\hline
\textcolor{blue}{\cite{masek2003matlab}} & \hspace{-1.4cm}IrisCode using 1D
  & \hspace{-0.4cm}(1) PolyU bi-spectral 
& \hspace{0.2cm}Genuine = 2800
& Cross-spectral &Hand-crafted & (1) 52.6\%
\\
&\hspace{-1.6cm}Log-Gabor filter & \hspace{-1.5cm}iris database & \hspace{1cm}Impostor = 1,953,000 &&&  (FAR = 0.1)\\
&&\hspace{-0.4cm}(2) Cross-eyed-cross- & \hspace{0.2cm}Genuine = 2160 &&&(2) 70.3\%\\
&& spectral iris recognition &\hspace{0.9cm}Impostors = 516,240 &&& (FAR = 0.1)\\
&&\hspace{-2cm}database\\
\hline
\textcolor{blue}{\cite{nalla2016toward}} & \hspace{-1.4cm}NIR to VIS texture
  & \hspace{-0.4cm}(1) PolyU bi-spectral 
& \hspace{0.2cm}Genuine = 2800
& Cross-spectral &Hand-crafted & (1) 64.91\%
\\
&\hspace{-1.1cm}synthesis using MRF & \hspace{-1.5cm}iris database & \hspace{1cm}Impostor = 1,953,000 &&&  (FAR = 0.1)\\
&\hspace{-3.1cm}model &\hspace{-0.4cm}(2) Cross-eyed-cross- & \hspace{0.2cm}Genuine = 2160 &&&(2) 78.13\%\\
&& spectral iris recognition &\hspace{0.9cm}Impostors = 516,240 &&& (FAR = 0.1)\\
&&\hspace{-2cm}database\\
\hline

\textcolor{blue}{\cite{wang2019cross}} & \hspace{-1.4cm}CNN with softmax
  & \hspace{-0.4cm}(1) PolyU bi-spectral 
& \hspace{0.2cm}Genuine = 2800
& Cross-spectral &self-learned & (1) 90.71\%
\\
&\hspace{-1.1cm}cross-entropy loss for & \hspace{-1.5cm}iris database & \hspace{1cm}Impostor = 1,953,000 &&&  (FAR = 0.01)\\
&\hspace{-1.2cm} feature extraction and  &\hspace{-0.4cm}(2) Cross-eyed-cross- & \hspace{0.2cm}Genuine = 2160 &&&(2) 87.18\%\\
&\hspace{-1.1cm}SDH for compression & spectral iris recognition &\hspace{0.9cm}Impostors = 516,240 &&& (FAR = 0.01)\\
&\hspace{-1.7cm}and classification&\hspace{-2cm}database\\
\hline
\textcolor{blue}{cpGAN}\cite{mostofa2020cross}&\hspace{-1.3cm}Conditional coupled&\hspace{-0.5cm}(1) PolyU bi-spectral&\hspace{0.4cm}Genuine = 2800&Cross-spectral&self-learned&(1) \textbf{92.38\%}\\
&\hspace{-1cm}generative adversarial&\hspace{-1.4cm}iris database&\hspace{1.2cm}Impostor = 1,953,000&and cross-&in the embedded &(FAR=0.01) \\
&\hspace{-1.5cm}network (cpGAN)&\hspace{0.2cm}
\hspace{0.2cm}&&resolution&domain\\
\hline
\end{tabular}}
\end{center}
\end{table*}

In the last few years, deep neural network architectures, such as a convolutional neural network (CNN), have dramatically improved the capabilities in automatically learning the deep representation of specific image features for object detection and classification of visual patterns. These algorithms have also shown superior results when compared to classical techniques based on hand-crafted features. Recently, successful deployment of deep learning architectures for the task of the same or cross-domain iris recognition has gained attention in the literature. Generative adversarial networks (GANs)\textcolor{blue}{\cite{goodfellow2014generative}}, among other deep neural network architectures, have shown outstanding capabilities in different areas of computer vision and biometric applications \textcolor{blue}{\cite{zhang2018tv,parkhi2015deep,cao2017vggface2,zhao2017towards,nguyen2017iris,minaee2019deepiris,radford2015unsupervised,bezerra2018robust,wang2018generative, minaee2018finger,minaee2018iris}}. A range of applications of GANs for iris recognition has been presented, including data augmentation, synthesis of NIR periocular images, synthesizing iris images and iris super-resolution\textcolor{blue}{\cite{lee2019conditional, tapia2019soft}}. In this paper, our main contribution is the extensive application of our novel algorithms on three publicly available iris datasets comparing two different GAN-based frameworks for cross-spectral (VIS vs NIR) and cross-resolution (low-resolution (LR) NIR to high-resolution (HR) VIS) iris matching, which resulted in a new state-of-the-art approach in the area of ocular biometrics.

We have developed two approaches by which we apply a family of deep learning frameworks for different cross-spectral iris matching scenarios. In our first approach, we employ a conditional GAN (cGAN)\textcolor{blue}{\cite{mirza2014conditional}} architecture to map the cross-spectral data to the same spectral domain. We apply it at the same resolution and extend it to the cross-resolution iris matching problem. We have designed our first method based on a scenario when one already has access to an Open-source or a commercial off-the-shelf iris matcher (e.g., Open-Source OSIRIS\textcolor{blue}{\cite{othman2016osiris}} matcher) to conduct the iris verification process. The key idea in our first method is to synthesize the VIS iris images from their corresponding NIR iris images in a gallery at the same resolution or higher resolution through a joint cross-modal super-resolution process. Our first method is assumed to be a preprocessing module that translates a NIR image into its corresponding VIS iris image before using a commercial iris matcher. In our work, we have used OSIRIS software to conduct the matching between the synthesized VIS iris images from a gallery of NIR iris images and a prob VIS iris image. In a summary, our first approach offers four contributions to the field of iris recognition:\vspace{0.1cm}\\
\indent • A new domain adaptation framework, which acts as a preprocessing module for cross-spectral iris matching based on generative adversarial networks to transform the cross-domain problem to the same domain and achieves comparable performance when compared to several state-of-the-art methods.\vspace{0.1cm}\\
\indent• Integrating the cross-resolution matching scenario into the cross-spectrum setting and redefining the matching framework as a joint super-resolution and cross-spectral matching architecture.\vspace{0.1cm}\\
\indent• Introducing a new WVU face and iris dataset, which will contribute to the biometric field for cross-spectral face and iris recognition.\vspace{0.1cm}\\
\indent• Performing substantial experiments on the PolyU Bi-Spectral dataset\textcolor{blue}{\cite{nalla2016toward,wang2019cross}}, WVU face and iris dataset and cross-eyed-cross-spectral iris recognition database\textcolor{blue}{\cite{sequeira2016cross}}.

We observe a significant improvement in the cross-spectral iris matching accuracy from the experimental results of our first approach, which validates that our domain adaptation technique requiring self-learned features extracted from the raw data can achieve remarkable performance gains for iris verification tasks similar to the previous research presented in the literature. However, it is still essential to explore a more compressed and distinctive representation of the raw data. In earlier works, researchers have widely used subspace learning for data representation\textcolor{blue}{\cite{lian2010max,ding2015missing,fu2008correlation}}. Basically, it has been proven that the most relevant and useful inner characteristics of an image can be mapped to a reduced low-dimensional latent subspace.\\  \indent Motivated by this, in our second method, we focus on the idea of designing a dedicated cross-spectral iris matcher completely avoiding the use of any commercial off-the-shelf iris matcher. We hypothesize that iris images in the VIS domain possess a latent connection with iris images in the NIR domain in a latent feature subspace. Therefore, we exploited this latent connection by projecting the VIS and NIR iris images into a common latent embedding subspace, even if they are at different resolutions. Furthermore, we posit that, if we perform matching in this latent domain, verification performance would be more accurate than our first method due to the inherent direct iris matching in a shared common feature domain. Moreover, our second method is designed to circumvent several shortcomings of the commercial iris matchers that our first method depends on. The idea can be elaborated on with a use of a case scenario.  For example, the matching engine of a commercial iris matcher cannot be adjusted to the resolution of the iris images captured at a distance. It cannot operate directly on the lower resolution images while enrolled images in the gallery, are comparatively, at a higher resolution. In addition, effective usage of opportunistic high-resolution VIS iris images captured by smartphones, surveillance cameras, etc. requires an accurate, fast, stable, and secure iris matcher. This can be achieved with the real-valued feature representation in the common embedded latent subspace instead of a binarized feature representation like the one used in other classical iris recognition approaches in the literature. These underlying reasons motivated us to develop such a dedicated cross-spectral iris matcher, which is highly desirable.  Hence,  we  have  proposed  a  deep  coupled  learning framework  for  cross-spectral  iris  recognition,  which  utilizes  a conditional coupled generative adversarial network (cpGAN)\textcolor{blue}{\cite{mostofa2020cross}} to  learn  a  common  embedded  feature  vector  via  exploring  the correlation  between  the  NIR and  VIS  iris  images  in  a  reduced dimensional latent embedding feature subspace. The key benefits from our second iris recognition approach can be summarized as the following:\vspace{0.2cm}\\
\indent • We analyze an effective method to learn the subspace embedded features and develop a novel framework for cross-spectral iris matching using our cpGAN architecture.\vspace{0.1cm}\\
\indent \hspace{0.01cm}• Comprehensive experiments on three different benchmark datasets (1) PolyU Bi-Spectral dataset (2) WVU face and Iris dataset and (3) Cross-eyed-cross-spectral database with superior results over the baseline approaches ascertain the validity of our cpGAN framework.\vspace{0.1cm}\\
\indent • To the best of our knowledge, this is the first study that has investigated two different techniques utilizing the potential capabilities of a GAN to improve the performance of existing cross-spectral iris recognition methods reported in the literature.

\begin{figure*}
\centering
\includegraphics[width=15.5cm]{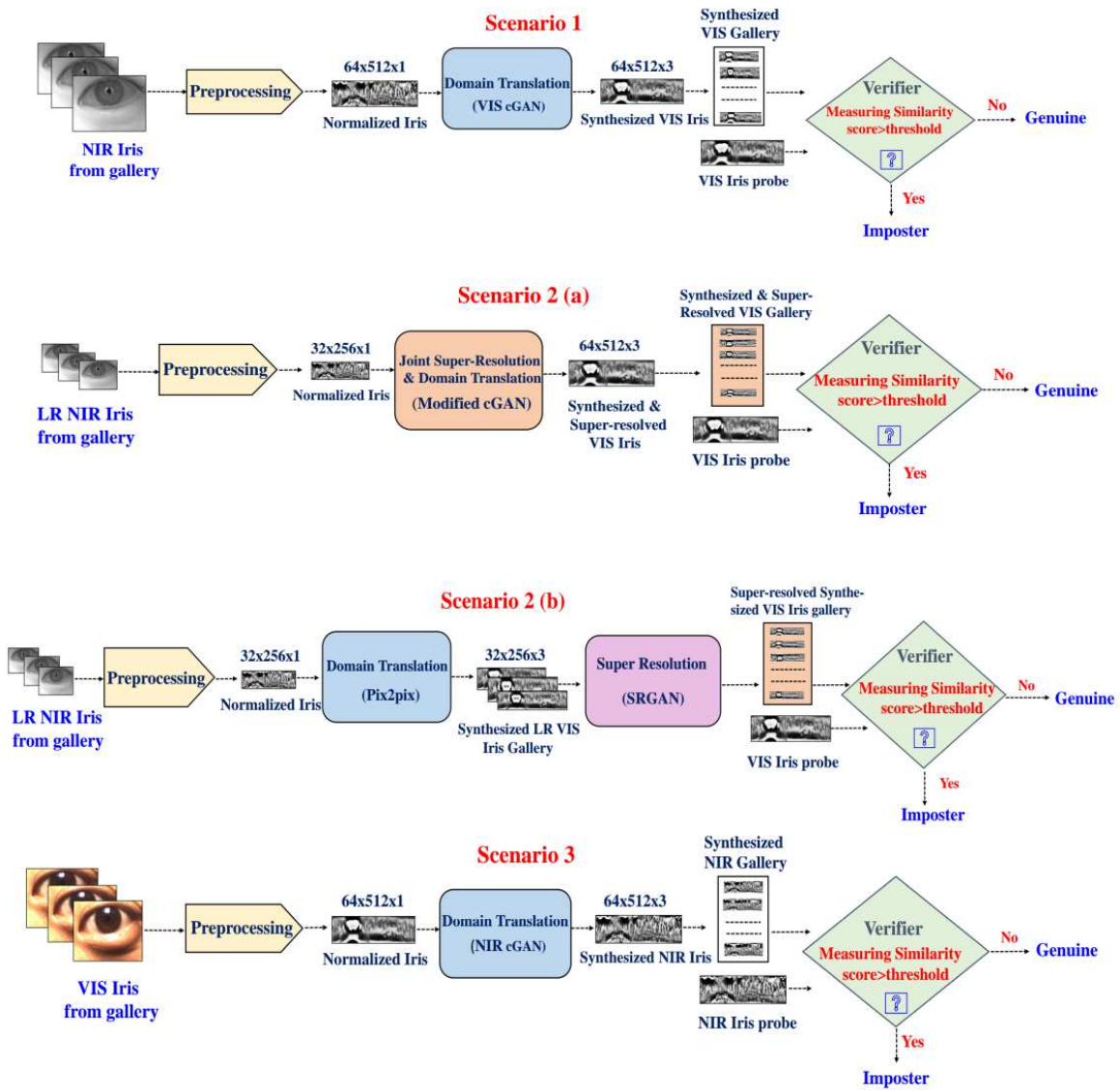}
\caption{Cross-domain and cross-resolution iris recognition framework; Scenario 1: NIR to visible translation; Scenario 2: NIR to visible
joint/separate translation and Super-resolution , Scenario 3: Visible to NIR translation.}
\end{figure*}

\section{Related Work}
Cross-spectral iris recognition requires a VIS iris probe to be matched against a
gallery of NIR iris images. While conventional iris recognition methods have achieved high matching accuracy, cross-spectral iris matching algorithms have not yet reached a high level of performance and pose a greater challenge for real-world applications.

\textcolor{blue}{Table 1} summarizes recent cross-spectral iris algorithms that are based on two strategies 1) extracting information from both spectral domains and then combining such information for the final decision, or 2) synthesizing a NIR image from its corresponding visible image and then matching against a NIR gallery. Using the first strategy, Vyas and Kanumuri\textcolor{blue}{\cite{vyas2019cross}} proposed a new feature descriptor using template partitioning based on variations in the iris texture. In their work, they have applied a 2D Gabor filter bank to obtain the iris pattern at various scales and orientations. They utilize the difference of variance (DoV) features to divide the filtered iris template into sub-blocks, as the DoV features are invariant to noise caused by illumination occlusion and position shifting. However, this method could not achieve the high accuracy required for practical applications (high equal error rate (EER) of 31.08\%) because it is unable to relate the information comprised in the NIR and VIS images. Tan et al. \textcolor{blue}{\cite{tan2012unified}} describe a framework for segmenting iris images in both domains which is helpful for further multi-spectral fusion of information. According to Oktiana et al.\textcolor{blue}{\cite{oktiana2018features}} local binary pattern (LBP) and binary statistical image feature (BSIF) are the best feature descriptors based on the VIS and NIR imaging systems, which are able to accurately extract the texture patterns of the iris for cross-spectral matching. 

Another recent work\textcolor{blue}{\cite{raja2016cross}}  also used BSIF along with the $\chi^{2}$ distance metric to obtain match scores between the VIS prob and NIR reference ocular images. They then fuse all the scores to make the final decision. To encourage advances in cross-spectral iris and periocular recognition, there has been a competition\textcolor{blue}{\cite{sequeira2017cross}} held among five teams, which is considered as an extension of 1st competition  that was arranged for a similar task (\textit{more recently, Sequeira et al.\textcolor{blue}{\cite{sequeira2016cross}} released a new cross-eyed and cross-spectral iris dataset to advance research on the challenging cross-spectral iris matching problem}). They submitted twelve methods for the periocular task and five for the iris task. In the work of Alonso-Fernandez et al.\textcolor{blue}{\cite{alonso2015near}} fusion of periocular and iris information achieved considerable match performance improvement, where iris information is obtained by using different iris features extraction techniques. Wild et al.\textcolor{blue}{\cite{wild2015fusion}} used information from iris images captured at multiple bands of the electromagnetic spectrum and presents an efficient feature-level fusion to improve cross-spectral iris recognition performance. Sharma et al.\textcolor{blue}{\cite{sharma2014cross}}  proposed an algorithm, that consists of two neural network architectures, and trained it on a cross spectral periocular dataset. It resulted in an improved matching accuracy compared to the existing feature descriptors previously mentioned above. 

On the other hand, using the second strategy, several efforts toward estimating NIR images from visible images have been proposed recently. For instance, researchers in\textcolor{blue}{\cite{zuo2010cross}} have explored an adaptive learning method to predict NIR images to address the performance shortcomings, which was considered below the benchmarks caused by cross-spectral matching. Similarly, in\textcolor{blue}{\cite{nalla2016toward}}, authors develop a domain adaption framework using Markov random fields (MRF) to estimate a NIR iris image from its corresponding VIS iris image and perform matching against a NIR image gallery. In the same direction, Burge and Monaco\textcolor{blue}{\cite{burge2009multispectral, burge2013multispectral}} implemented a model which utilized features derived from the color and structure of the VIS iris images to predict the corresponding synthesized NIR iris images. We have also noticed similar works in the ocular biometric field for the task of cross-spectral periocular image recognition. Recently, Reja et al.\textcolor{blue}{\cite{raja2019cross}} proposed a novel image transformation technique using cascaded refinement networks to synthesize a NIR periocular image from the corresponding VIS periocular image. Another study\textcolor{blue}{\cite{oktiana2020cross}} reported that feature-based approaches are prone to changes during the feature extraction process. Therefore, they have adopted phase-only correlation and band-limited phase-only correlation techniques to develop a phase-based iris recognition system. 

Although the approaches mentioned above have advanced cross-spectral iris matching one step ahead by achieving good results, but to keep pace with the increasing demand for more robust biometric systems, researchers have recently concentrated their efforts towards CNN-based iris verification system\textcolor{blue}{\cite{wang2019cross}}. In this study, the authors observed that CNN-based features offer a significantly compact representation for the iris template along with sparse information, which potentially helps to improve the accuracy of the iris recognition system. Moreover, this approach incorporates a supervised discrete hashing (SDH) on the learned features, which achieved an EER of 5.39\%.\vspace{0.2cm}\\ 
\indent Another interesting approach, iris image super-resolution, has also gained attention due to its impact on iris verification methods. The authors in\textcolor{blue}{\cite{ribeiro2017exploring}} explored deep learning architectures such as stacked auto-encoders and CNN for single-image iris super-resolution. Wang et al.\textcolor{blue}{\cite{wang2019iris}} proposed a framework based on an adversarial training with triplet networks in order to improve iris image resolution for further recognition.

\begin{figure*}
\centering
\includegraphics[width=14 cm]{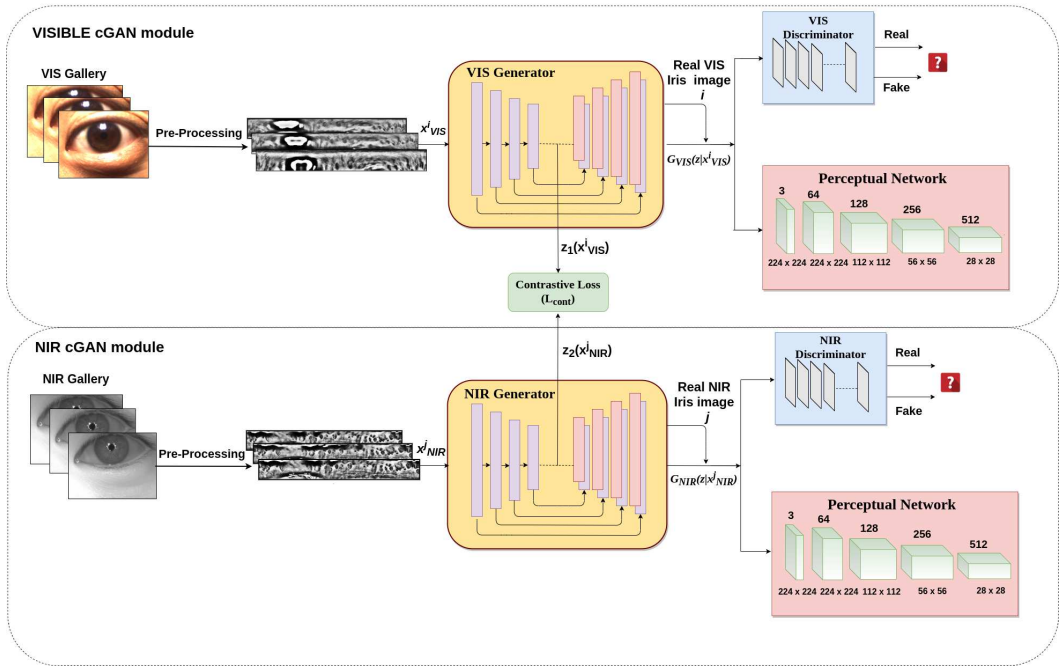}
\caption{Architecture of our proposed conditional cpGAN framework. During training, the contrastive loss function is used in the latent embedding subspace to optimize the network parameters so that latent features of iris images from different spectral domain of the same identity are close to each other while the features of different identities are pushed further apart\textcolor{blue}{\cite{mostofa2020cross}}.}
\end{figure*}

\section{Methodology}
To address the performance degradation reported in cross-spectral iris matching, our primary goal is to develop an algorithm that minimizes the distance between the VIS iris image and NIR iris image distributions belonging to the same person. Therefore, we have developed two different domain adaptation techniques. These two techniques are based on finding a mapping, or a low-dimensional shared latent subspace, between the VIS and NIR iris modalities to significantly reduce the cross-spectral iris matching discrepancy and provide a new state-of-the-art result. The techniques developed in this paper are shown in \textcolor{blue}{Fig. 1} and \textcolor{blue}{Fig. 2}, respectively. Since most of the available iris galleries are acquired under NIR illumination and the opportunistic iris images are obtained under the VIS domain at higher resolution, in our first technique we find a mapping between the NIR and VIS iris images and design a new framework (see \textcolor{blue}{Fig. 1}) based on joint cross-domain and cross-resolution matching to enable cross-spectrum iris recognition for pairs of images with the same and different resolutions. In greater detail, we address two challenges: 1) performing cross-domain mapping for the purpose of the intra-domain iris matching, and 2) doing the same when the images from each spectral domain have different resolutions. As a result, there should be a joint transformation of spectrum and resolution, which will be discussed in detail in subsections 3.1.1 and 3.1.2.

We consider three main scenarios, as shown in \textcolor{blue}{Fig. 1}, and develop our iris matching framework around them based on different cross-spectral scenarios. The first scenario is matching a visible probe against an NIR gallery translated to an equivalent visible gallery. The second scenario is matching a high-resolution visible probe against a translated and super-resolved NIR gallery to a high-resolution visible gallery. The third scenario is matching an NIR probe against a visible gallery translated to an equivalent NIR gallery. The reason behind including the cross-resolution setting in our framework, is that, as an emerging problem domain, current opportunistic visible iris images extracted from high-resolution face images are typically at a higher resolution than the NIR images.

Apart from being focused solely on the generation of a synthesized VIS image from its NIR counterpart, in our second approach, we emphasize the idea of learning a latent subspace to extract meaningful representative features from the VIS and NIR iris images. Thus, we develop our second approach as shown in \textcolor{blue}{Fig. 2}, which projects both the NIR and VIS iris images to a common latent low-dimensional embedding subspace using two generative networks. The key reason behind developing this architecture is to learn the semantic similarity between two samples of the same subject but in different spectral domains. Therefore, inspired by our previous cpGAN architecture \textcolor{blue}{\cite{mostofa2020cross}}, we trained this network using a similarity measure based on a contrastive loss\textcolor{blue}{\cite{chopra2005learning}} to ensure that the distance between the images corresponding to the genuine pairs (VIS iris image and NIR iris image of the same person) is minimized, and that of the imposter pairs (VIS iris image and NIR iris image of the different persons) is maximized.

To summarize our two approaches, we have studied and developed two different deep convolutional GAN-based architectures to ascertain the adaptive learning potential for cross-spectral iris matching, i.e., cGAN and cpGAN. Like other approaches, before training both networks, preprocessing steps require iris images from both spectra to be subjected to segmentation, normalization and image enhancement processes. Then, normalized image samples are fed to each network. The network in our first approach mainly utilizes the adversarial loss to synthesize VIS iris image from its NIR counterpart before performing different verification scenarios, while in our second approach, the network integrates the contrastive loss along with the adversarial learning\textcolor{blue}{\cite{goodfellow2014generative}} to generate matching scores. The following sections provide the details of our approaches and introduce the cGAN and cpGAN architectures along with the associated loss functions that are implemented in our framework to investigate the cross-spectral iris matching problem.

\subsection{Deep Conditional  Adversarial Framework}
Recently, GANs have received considerable attention from the deep learning research community due to their significant contributions in the field of image generation. The basic GAN framework consists of two modules$-$ a generator module, G, and a discriminator module, D. The objective of the generator, G, is to learn a mapping, $G:z\rightarrow y$, so that it can produce synthesized samples from a noise variable, $z$, with a prior noise distribution, $p_{z}(z)$, which is difficult for the discriminator, D, to distinguish from the real data distribution, $p_{data}$, over $y$. The generator, $G(z; \theta_{g})$ is a differentiable function which is trained with parameters $\theta_{g}$ when mapping the noise variable, $z$, to the actual data space, $y$. Simultaneously, the discriminator, D, is trained as a binary classifier with parameters $\theta_{d}$ such that it can distinguish the real samples, $y$, from the fake  ones, $G(z)$. Both the generator and discriminator networks compete with each other in a two-player minimax game. We calculate the following loss function, $L(D,G)$, for the GAN:
\begin{equation}\begin{split}
\centering
     L(D,G) & = E_{y\sim P_{data}(y)}[\log D(y)]\\ & + E_{z\sim P_{z}(z)}[\log (1-D(G(z)))]. \end{split}
 \end{equation}

The objective function of GAN defines the term “two-player minimax game” by optimizing the loss function, $L(D,G)$, as follows: 
\begin{equation}\begin{split}
     \min_{G}\max_{D} L(D,G) & =\min_{G}\max_{D}[E_{y\sim P_{data}(y)}[\log D(y)]\\ & + E_{z\sim P_{z}(z)}[\log (1-D(G(z)))]].\end{split}\label{eq:2}
\end{equation}

One of the variants of GAN, the cGAN is introduced in \textcolor{blue}{\cite{mirza2014conditional}}, which expands the scope of synthesized image generation by setting a condition for both the generative and discriminative networks. The cGAN applies an auxiliary variable, $x$, as a condition which could be any kind of useful information such as texts \textcolor{blue}{\cite{reed2016generative}}, images \textcolor{blue}{\cite{isola2017image}} or discrete labels \textcolor{blue}{\cite{mirza2014conditional}}. The loss function for the cGAN, $L_{c}(D,G)$, can be represented as follows: 
\begin{equation}\begin{split}
     L_{c}(D,G) & = E_{y\sim P_{data}(y)}[\log D(y|x)]\\ & + E_{z\sim P_{z}(z)}[\log (1-D(G(z|x)))].\end{split}\label{eq:3}
\end{equation} Similar to (2), the objective function of the cGAN is minimized in a two-player minimax manner, which is denoted as $L_{cGAN}(D,G,y,x)$ and defined by: \begin{equation}\begin{split}
L_{cGAN}(D,G,y,x) & = \min_{G}\max_{D} [E_{y\sim P_{data}(y)}[\log D(y|x)]\\ & + E_{z\sim P_{z}(z)}[\log (1-D(G(z|x)))]].\end{split}\label{eq:4}
\end{equation}

\subsubsection{Domain Translation Using cGAN}

A more recent algorithm in the field of ocular biometrics\textcolor{blue}{\cite{nalla2016toward,ramaiah2016advancing}} has shown success in estimating NIR iris images from VIS iris images and then matching them against the NIR instances in the gallery. However, they did not use CNN-based algorithms, even though many of the recent iris recognition systems have investigated the capabilities of CNN in learning anatomical properties. Therefore, we have developed a deep CNN-based domain translation network in our first method. We proposed to translate the iris images from the NIR domain to visible, or vice versa. Therefore, image translation plays an important role as one of two integral parts of our frameworks.
 
Recent advances in deep learning reported in the literature have provided very powerful tools for the task of image-to-image translation\textcolor{blue}{\cite{isola2017image}}. Such translations can be interpreted as image domain transformations, where the task is to learn a mapping from one modality to another modality.  In our first method, we use the conditional GAN (cGAN) architecture\textcolor{blue}{\cite{isola2017image}} for the task of NIR to VIS iris image translation or vice versa.  The cGAN architecture has been successful in a variety of image-to-image translation tasks in the computer vision research community. It includes Sketch → Portrait, Sketch→ Pokemon, Depth→ Streetview, pose transfer, etc. Such deployment of cGAN in image translation tasks has inspired us to explore its performance in synthesizing corresponding VIS iris images from the NIR iris gallery, to be used as a preprocessing module for the cross-spectral iris image translation.


During training the cGAN, we condition on an NIR iris image and generate a corresponding synthesized VIS iris output image or vice versa as shown in our proposed framework (see \textcolor{blue}{Scenario 1} and \textcolor{blue}{Scenario 3} in \textcolor{blue}{Fig. 1}). Here, we have demonstrated that a simplified cGAN framework is sufficient to achieve adequate synthesized results through adversarial learning. In addition, our analysis shows that this method is effective at conducting cross-spectral iris matching under the same spectrum setting (a VIS iris probe is matched against a synthesized VIS gallery generated from its corresponding NIR gallery or vice versa.) with impressive results.

\subsubsection{Joint Translation and Super-Resolution Using Modified cGAN}
Leveraging the benefits of the cGAN architecture, we have investigated the possibility of iris domain translation  by using a structured loss\textcolor{blue}{\cite{isola2017image}} to penalize any probable structural mismatch between the synthesized output and target. Successful deployment of this network helps us to overcome the challenge faced in cross-spectral iris matching. However, in \textcolor{blue}{Scenarios 2(a)} and \textcolor{blue}{2(b)}, representing the additional cross-resolution case (see \textcolor{blue}{Fig. 1}), the size of the output image should be larger than the size of the input image, i.e., the network should learn domain translation to a higher resolution. In this context, we modify the architecture of our cGAN generator by integrating the concept of super-resolution during the cross-domain translation. Super-Resolution (SR) estimates a HR super-resolved image from its LR counterpart, which has been vigorously applied to various computer vision applications. Although reconstructing an accurate HR image from its LR version is a very difficult task, multiple SR algorithms have been developed in recent years\textcolor{blue}{\cite{wang2020deep}} to address this challenge. 

Recently, the GAN-based SRGAN\textcolor{blue}{\cite{ledig2017photo}} approach has shown excellent results with high perceptual image quality by retrieving the fine textural details from a LR input image. Following their approach of up-sampling the LR input image, we improve our cGAN-based translation architecture and incorporate a super-resolution layer as part of our cross-spectral framework to deal with the cross-resolution task considered in our basic  \textcolor{blue}{Scenario 2}. To synthesize high-quality VIS iris images, we train our network with a perceptual loss\textcolor{blue}{\cite{johnson2016perceptual}}, which helps to generate a more accurate VIS iris images along with the widely used $L_{2}$ reconstruction loss\textcolor{blue}{\cite{dong2015image, shi2016real}} and the adversarial loss\textcolor{blue}{\cite{ledig2017photo}} functions. 
A similar iris super-resolution method has been proposed in\textcolor{blue}{\cite{wang2019iris}}, which integrates 
adversarial training into triplet networks in order to develop a super-resolution architecture for low-quality iris images.
However, the ability of their SR network is limited to super-resolving iris images in the same spectral domain. On the other hand, we jointly perform super-resolution and domain transformation in one shot
to overcome the limitations of acquiring high-resolution NIR iris images. More specifically, our network produces a gallery of super-resolved HR VIS iris images from a gallery of LR NIR iris images, which is then used to match a HR VIS iris probe against it. 

\subsection{Verification}
In this article, we have proposed to perform cross-spectral iris matching under the same spectra and the same resolution setting by adopting joint translation and super-resolution technique followed by the verification process. To accomplish this, we train our network on unrolled iris images of one spectral domain as input and generate unrolled iris images of the other spectral domain at the same resolution or higher resolution based on the scenarios described in the earlier sections. To perform verification, we employ a commercially available software, Open Source for IRIS (OSIRIS), which was developed within the BioSecure project\textcolor{blue}{\cite{othman2016osiris}} and offered by its authors as a free, open-source iris matcher. OSIRIS follows the iris matching concept proposed in the works of Daugman\textcolor{blue}{\cite{daugman2009iris}}. It applies Daugman’s rubber sheet model for unwrapping the iris image from polar coordinates onto a Cartesian rectangle to process image segmentation and normalization tasks. Hence, during verification we match a normalized VIS iris probe against a gallery of synthesized normalized VIS iris images generated from our network using this OSIRIS software. It first generates iris codes by applying phase quantization of multiple Gabor wavelet filtering outcomes, while matching is performed using XOR operation, with normalized Hamming distance as an output dissimilarity metric. For genuine comparisons, we expect values close to zero, while we expect scores around 0.5 for imposter comparisons.

\subsection{Deep Coupled Adversarial Framework}

Our second proposed technique is a cpGAN architecture that consists of two coupled cGAN modules with the same architecture, as shown in \textcolor{blue}{Fig. 2}. One of them is dedicated to synthesizing the VIS iris images, and hence, we refer to as the VIS cGAN module. Similarly, the other module is dedicated to synthesizing the NIR iris images, which is referred to as the NIR cGAN module. Our cpGAN network is inspired by the Siamese network\textcolor{blue}{\cite{bromley1993signature}}, which ensures pairwise learning, where all the parameters are simultaneously updated throughout the network. We have followed a more recent U-Net-based, densely-connected  encoder-decoder structure proposed in\textcolor{blue}{\cite{zhang2018densely}} to design our generator, which helps to achieve the low-dimensional embedded subspace for cross-spectral iris matching via a contrastive loss along with the standard adversarial loss. In addition to the adversarial loss and contrastive loss\textcolor{blue}{\cite{chopra2005learning}}, the perceptual loss\textcolor{blue}{\cite{johnson2016perceptual}}, and $L_{2}$ reconstruction loss are also used to guide the generators towards the optimal solutions. Perceptual loss is measured via a pre-trained VGG 16 network\textcolor{blue}{\cite{simonyan2014very}}, which helps in sharp and realistic reconstruction of the images. In realistic opportunistic iris recognition scenarios, a VIS iris probe is usually matched against a gallery of NIR iris images. To create such application scenario, we focus on matching a VIS iris probe against a gallery of NIR iris images, that have not been seen before by the network. To perform this matching in a cross-spectral domain setting, a discriminative model is required to produce a domain-invariant representation. Therefore, we force the network to learn iris feature representations in a common embedding subspace by utilizing a U-Net auto-encoder architecture that uses class-specific contrastive loss to match the iris patterns in the latent domain. As previously mentioned, we use a U-Net auto-encoder architecture for our generator due to its structural ability of extracting features in the latent embedding subspace. More specifically, the contracting path of the “U shaped” structure of the U-Net captures contextual information, which is passed directly across all the layers, including the bottleneck. In neural networks, the bottleneck forces the network to learn the compressed version of the input data that only contains useful information to preserve the structural integrity of the image required to reconstruct the input. Along with the bottleneck, the high-dimensional features of the contracting path of the U-Net, combined with the corresponding upsampled features of the symmetric expanding path, provides a means to share the useful information throughout the network. Moreover, during domain transformation, a significant amount of low-level information needs to be shared between input and output, which can be accomplished by leveraging a U-Net-like architecture.

We have followed the architecture of patch-based discriminators\textcolor{blue}{\cite{isola2017image}} to design the discriminators of our proposed model. The discriminators are trained simultaneously along with the respective generators. It is worthwhile to mention that the $L_{1}$ loss performs very well when applied to preserve the low-frequency details but fails to preserve the high-frequency information, whereas patch-based discriminator penalizes the structure at the patch scale to ensure the preservation of high-frequency details.

The main idea behind using the U-Net shaped generator
is to gradually build a connection between the VIS and NIR iris images in the common embedding feature subspace. Since the features are domain invariant in the embedded subspace, it provides credibility to discriminate images based on identity. Therefore, our final objective is to find a set of domain invariant features in a common latent embedding subspace by coupling the two generators via a contrastive loss function, $L_{cont}$\textcolor{blue}{\cite{chopra2005learning}}.

The contrastive loss function,  $L_{cont}$, is defined as a distance-based loss metric, which is computed over a set of pairs in the common embedding subspace such that images belonging to the same identity (genuine pairs i.e., a VIS iris image of a subject with its corresponding NIR iris image) are embedded as close as possible, and images of different identities (imposter pairs i.e., a VIS iris image of a subject with a NIR iris image of a different subject) are pushed further apart from each other. The contrastive loss function is formulated as:
\begin{equation}
\begin{split}
L_{cont}(z_1&(x^i_{VIS}),z_2(x^j_{NIR}),Y)= \\ & 
  (1-Y)\frac{1}{2}(D_z)^2 + (Y)\frac{1}{2}(\mbox{max}(0,m-D_z))^2,  
  \end{split}
  \end{equation}where
$x^i_{VIS}$ and $x^j_{NIR}$ denote the input VIS and NIR iris images, respectively. The variable, $Y$, is a binary label, which is set to 0 if $x^i_{VIS}$ and $x^j_{NIR}$ belong to the same class (i.e., genuine pair), and equal to 1 if $x^i_{VIS}$ and $x^j_{NIR}$ belong to different classes (i.e., impostor pair). $z_1(.)$ and $z_2(.)$ are denoted as the encoding functions of the U-Net auto-encoder, which transform  both $x^i_{VIS}$ and $x^j_{NIR}$, respectively into a common latent embedding subspace. Here, $m$, is used as the contrastive margin to ``tighten" the constraint. The Euclidean distance, $D_z$, between the outputs of the functions, $z_1(x^i_{VIS})$, and $z_2(x^j_{NIR})$, is given by:
 
\begin{equation}
     D_z=\left\lVert z_1(x^i_{VIS})-z_2(x^j_{NIR})\right\rVert_2.
\end{equation}
 
 Therefore, if $Y=0$ (i.e., genuine pair), then the contrastive loss function, $(L_{cont})$, is given as:
 \begin{equation}
{L_{cont}(z_1(x^i_{VIS}),z_2(x^j_{NIR}),Y)  = \frac{1}{2}\left\lVert z_1(x^i_{VIS})-z_2(x^j_{NIR})\right\rVert^2_2}, 
\end{equation} and if $Y=1$ (i.e., impostor pair), then the contrastive loss function, $(L_{cont})$, is :
  \begin{equation}
  \begin{split}
L_{cont}(z_1(x^i_{VIS}),&z_2(x^j_{NIR}),Y)  = \\ & \frac{1}{2}\mbox{max}\biggl(0,m-\left\lVert z_1(x^i_{VIS})-z_2(x^j_{NIR})\right\rVert^2_2\biggr).
\end{split}
\end{equation}

Thus, the total loss  for coupling the VIS generator and NIR generator is denoted by $L_{cpl}$ and is given as:
\vspace{-0.25cm}
\begin{equation}
    \begin{split}
        L_{cpl}=\frac{1}{N^2}\sum_{i=1}^{N}\sum_{j=1}^{N}L_{cont}(z_1(x^i_{VIS}),z_2(x^j_{NIR}),Y), 
    \end{split}\label{eq:6}
\end{equation}
where N is the number of training samples. The contrastive loss in the above equation can also be replaced by some other distance-based metric, such as the Euclidean distance. However, the main aim of using the contrastive loss is to be able to use the class labels implicitly and find a discriminative embedding subspace, which may not be the case with some other metric such as the Euclidean distance. This discriminative embedding subspace would be useful for matching the VIS iris images against the gallery of NIR iris images.

\section{Loss Functions}
\subsection{Generative Adversarial Loss}
The VIS and NIR generators are denoted as $G_{VIS}$ and $G_{NIR}$, as they will synthesize the corresponding VIS and NIR iris images from the input VIS and NIR iris images, respectively. The patch-based discriminators used for the VIS and NIR iris GANs are denoted as $D_{VIS}$ and $D_{NIR}$, respectively. To implement our proposed methods, we have used the conditional GAN, where the generator networks $G_{VIS}$ and $G_{NIR}$ are conditioned on the input VIS and NIR iris images, respectively. In addition, we have trained the generators and the corresponding discriminators with the  cGAN loss function \cite{mirza2014conditional} to ensure a real-looking natural image reconstruction such that the discriminators cannot distinguish the generated images from the real ones. Let $L_{VIS}$ and $L_{NIR}$ denote the cGAN loss functions for the VIS and NIR GANs, respectively. Therefore, the loss function for the cGAN which is considered as the backbone architecture in our first approach, can be defined as following: 
\vspace{-0.2cm}
\begin{equation}
L_{VIS}=L_{cGAN}(D_{VIS},G_{VIS},y^i_{VIS},x^i_{VIS}),
\end{equation}
\begin{equation}
L_{NIR}=L_{cGAN}(D_{NIR},G_{NIR},y^j_{NIR},x^j_{NIR}), 
\end{equation}where $L_{cGAN}$ is defined as the cGAN objective function in (\ref{eq:4}). The term, $x^i_{VIS}$, is used to denote the VIS iris image, which is defined as a condition for the VIS cGAN, and $y^i_{VIS}$,  is denoted as the real VIS iris image. It is worth mentioning that the real VIS iris image, $y^i_{VIS}$, is same as the network condition given by $x^i_{VIS}$. Similarly, $x^j_{NIR}$, denotes the NIR iris image that is used as a condition for the NIR cGAN. Again, like  $y^i_{VIS}$, the real NIR iris image, $y^j_{NIR}$, is same as the network condition given by $x^j_{NIR}$. The total adversarial loss for our proposed cpGAN is given by: 
\begin{equation}
L_{GAN}=L_{VIS}+L_{NIR}.
\end{equation}

\subsection{\texorpdfstring{$L_{2}$}{Reconstruction Loss} %
Reconstruction Loss}
We consider the $L_2$ reconstruction loss as a classical constraint for both the VIS cGAN and NIR cGAN to ensure better results. The $L_2$ reconstruction loss measures the reconstruction error in terms of the Euclidean distance between the reconstructed iris image and the corresponding real iris image.
We denote the reconstruction loss for the VIS cGAN as $L_{2_{VIS}}$ and define it as:
\begin{equation}
    L_{2_{VIS}}=\left\lVert G_{VIS}(z|x^i_{VIS})-y^i_{VIS}\right\rVert^2_2,
\end{equation}where $y^i_{VIS}$ is the ground truth VIS iris image, and $G_{VIS}(z|x^i_{VIS})$, is the output of the VIS generator.

Similarly, we denote the reconstruction loss for the NIR cGAN as $L_{2_{NIR}}$: 

\begin{equation}
    L_{2_{NIR}}=\left\lVert G_{NIR}(z|x^j_{NIR})-y^j_{NIR}\right\rVert^2_2,
\end{equation}where $y^j_{NIR}$ is the ground truth NIR iris image, and $G_{NIR}(z|x^j_{NIR})$, is the output of the NIR generator. Depending on the different cross-spectral iris matching scenarios, we use either $L_{2_{VIS}}$ or $L_{2_{NIR}}$ as the reconstruction loss, which is again  generally termed as $L_{2_{cGAN}}$ for the method proposed in our first approach.

For the cpGAN architecture proposed in our second approach, the  total $L_{2_{cpGAN}}$ reconstruction loss can be defined by the following equation:
\begin{equation}
    L_{2_{cpGAN}}=\frac{1}{N^2}\sum_{i=1}^{N}\sum_{j=1}^{N}(L_{2_{VIS}}+L_{2_{NIR}}).
\end{equation}
\subsection{Perceptual Loss}\label{subsec:percloss}
Although the GAN loss and the reconstruction loss are used to guide the generators, they fail to reconstruct perceptual features in the generated images. Perceptual features are defined by the visual characteristics of objects, which provide a perceptually pleasing look to the image. Hence, we have also used the perceptual loss, introduced in\textcolor{blue}{\cite{johnson2016perceptual}}, for style transfer and super-resolution. The perceptual loss function basically measures high-level differences, such as content and style dissimilarity, between images. The perceptual loss is based on high-level representations from a pre-trained VGG-16\textcolor{blue}{\cite{simonyan2014very}} like CNN. Moreover, it helps the network to generate better and sharper high-quality images \textcolor{blue}{\cite{johnson2016perceptual}}. As a result, it can be a significant alternative to solely using  the $L_1$ or $L_2$ reconstruction error. Recently, Zhang et al.\textcolor{blue}{\cite{zhang2018unreasonable}} introduced the LPIPS loss metric, which has been adapted in several deep learning architectures for image reconstruction. Therefore, it can be considered as an alternative loss function for perceptual fidelity instead of the well-known ImageNet pre-trained VGG-based perceptual loss\textcolor{blue}{\cite{johnson2016perceptual}}.  

In both of our approaches, we have added perceptual loss to both the VIS and NIR cGAN modules using a pre-trained VGG-16 network. It involves extracting the high-level features (ReLU3-3 layer) of VGG-16 for both the real input image and the reconstructed output of the generator. The perceptual loss calculates the $L_1$ distance between the features of real and reconstructed images to guide the generators $G_{VIS}$ and $G_{NIR}$. The perceptual loss for the VIS cGAN network is defined as:

\begin{equation}
    \begin{split}
        L_{P_{VIS}}=&\frac{1}{C_pW_pH_p}\sum_{c=1}^{C_{p}}\sum_{w=1}^{W_{p}}\sum_{h=1}^{H_{p}} \\ & {\left\lVert V(G_{VIS}(z|x^i_{VIS}))^{c,w,h}-V(y^i_{VIS})^{c,w,h}\right\rVert},
    \end{split}
\end{equation}
where $V(.)$ is used to denote a particular layer of the VGG-16 and $C_p$, $W_p$, and $H_p$ denote the layer dimensions.

Likewise, the perceptual loss for the NIR cGAN network is:

\begin{equation}
    \begin{split}
        L_{P_{NIR}}=&\frac{1}{C_pW_pH_p}\sum_{c=1}^{C_{p}}\sum_{w=1}^{W_{p}}\sum_{h=1}^{H_{p}} \\ & {\left\lVert V(G_{NIR}(z|x^j_{NIR}))^{c,w,h}-V(y^j_{NIR})^{c,w,h}\right\rVert}.
    \end{split}
\end{equation}Here, we simply define $L_{P_{VIS}}$ or $L_{P_{NIR}}$ as $L_{P_{cGAN}}$ to calculate perceptual loss for our first approach. The total perceptual loss function for the cpGAN  is given by:
\begin{equation}
    L_{P_{cpGAN}}=\frac{1}{N^2}\sum_{i=1}^{N}\sum_{j=1}^{N}(L_{P_{VIS}}+L_{P_{NIR}}).
\end{equation}

\begin{figure*}
\centering
\includegraphics[width=16 cm]{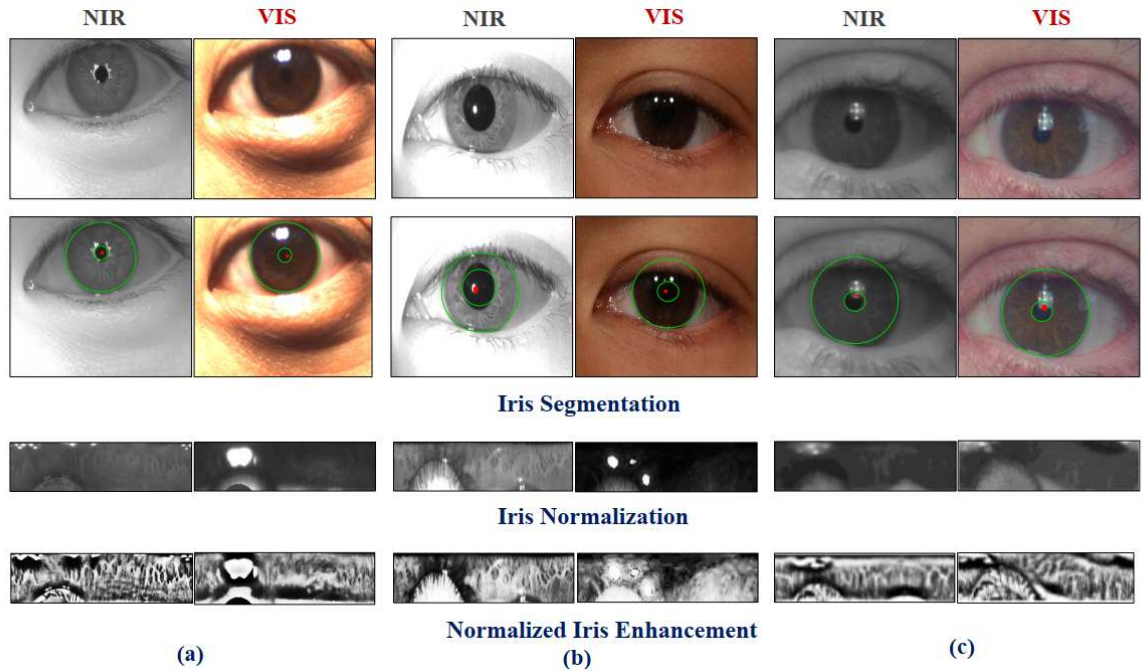}
\caption{Iris image preprocessing steps (Segmentation, Normalization, Enhancement) for (a) PolyU bi-spectral iris database and (b) WVU face and iris dataset (c) Cross-eyed-cross-spectral iris recognition database.}
\end{figure*}

\subsection{Overall Objective Function}
We sum all the loss functions defined above to calculate the overall objective function    $L_{tot_{cGAN}}$ and $L_{tot_{cpGAN}}$ for our proposed cGAN and cpGAN architectures, respectively: 
\vspace{-0.25cm}

\begin{equation}
    \begin{split}
      L_{tot_{cGAN}}= L_{2_{cGAN}} +\lambda_1 L_{cGAN}+  \lambda_2 L_{P_{cGAN}}
    \end{split}\label{eq:19},
\end{equation}where $L_{2_{cGAN}}$ is the total reconstruction
error, $L_{cGAN}$ is the total conditional generative adversarial loss function, and $L_{P_{cGAN}}$ is the total perceptual loss for our proposed cGAN model. Variables $\lambda_1$, and $\lambda_2$ are the adjustable hyper-parameters used to weigh the different loss terms.
The total loss for cpGAN is given as: 
\begin{equation}
    \begin{split}
       L_{tot_{cpGAN}}=L_{cpl}+ \lambda_3 L_{GAN} + \lambda_4 L_{P_{cpGAN}}+ \lambda_5 L_{2_{cpGAN}},
    \end{split}\label{eq:20}
\end{equation}where $L_{cpl}$ is the coupling loss, $L_{GAN}$ is the total generative adversarial loss, $L_{P_{cpGAN}}$ is the total perceptual loss, and $L_{2_{cpGAN}}$ is the total reconstruction error. Variables $\lambda_3$, $\lambda_4$, and $\lambda_5$ are the hyper-parameters used as a weight factor to numerically balance the magnitude of different loss terms. 
\vspace{-0.3cm}

\section{Experiments}
We first briefly introduce the publicly available datasets that we have used in our experiments and discuss the implementation details of our proposed cGAN and cpGAN architectures along with their training setup. To evaluate the performance of our methods, we perform a range of experiments for different cross-spectral iris matching scenarios and compare their performance with other state-of-the-art iris recognition methods in the  cross-domain setting. We provide detailed comparative experimental results in the following sections. Finally, in order to ascertain the usefulness of our cross-spectral iris recognition frameworks, we conduct additional experiments for cross-device iris matching scenarios.  
\subsection{Database}

Three available cross-spectral database, PolyU bi-spectral iris database\textcolor{blue}{\cite{nalla2016toward}}, WVU Face and Iris Dataset\footnote{\textcolor{red}{This data was collected at WVU under IRB \# 1805125982 with appropriate human subjects' approval.}},\footnote{\textcolor{red}{This dataset is available upon request at biic.wvu.edu.}} and Cross-eyed-cross-spectral iris recognition database\textcolor{blue}{\cite{sequeira2016cross}} are employed to validate our proposed methods.

\subsubsection{PolyU Bi-Spectral iris database}
The PolyU Bi-Spectral iris database contains iris images of 209 subjects acquired simultaneously under both the VIS and NIR illuminations. Each subject consists of 15 different instances of right and left-eye images with a resolution of $640 \times 480$ pixels for both VIS and NIR spectrum. Therefore, the total number of images in this dataset is 12,540 ($209\times2\times2\times15$). We used a publicly-available segmentation algorithm\textcolor{blue}{\cite{zhao2015accurate}} to accurately segment and normalize iris images for the experiments. This segmentation algorithm provides normalized iris images of $512\times64$ pixels, samples of which are shown in \textcolor{blue}{Fig. 3(a)}. Following the approach used in\textcolor{blue}{\cite{nalla2016toward}}, we selected the first ten instances for our network training and the remaining five instances for the testing. The all-to-all matching protocol generated 2,800 genuine scores and 1,953,000 imposter scores.

\subsubsection{WVU face and iris dataset}
The West Virginia University (WVU) Face and Iris dataset is particularly developed for cross-spectral opportunistic iris recognition. It contains 1,248 subjects, which provides a total of 2,496 left and right NIR as well as VIS iris images ($1,248\times2$).  We use the method presented in\textcolor{blue}{\cite{zhao2015accurate}} to extract the normalized iris images $(512\times64)$ from the original iris images of size $640\times480$ pixels. 
Sample images from this dataset are shown in \textcolor{blue}{Fig. 3(b)}.
Again, following the same train-test protocol used in reference\textcolor{blue}{\cite{nalla2016toward}} for this dataset, we attained 750 genuine scores and 561,750 imposter scores for 375 test subjects.
\begin{figure*}%
\vspace*{-0.4cm}   
\centering
\subfloat[]{\includegraphics[width=17cm,height=5.4cm]{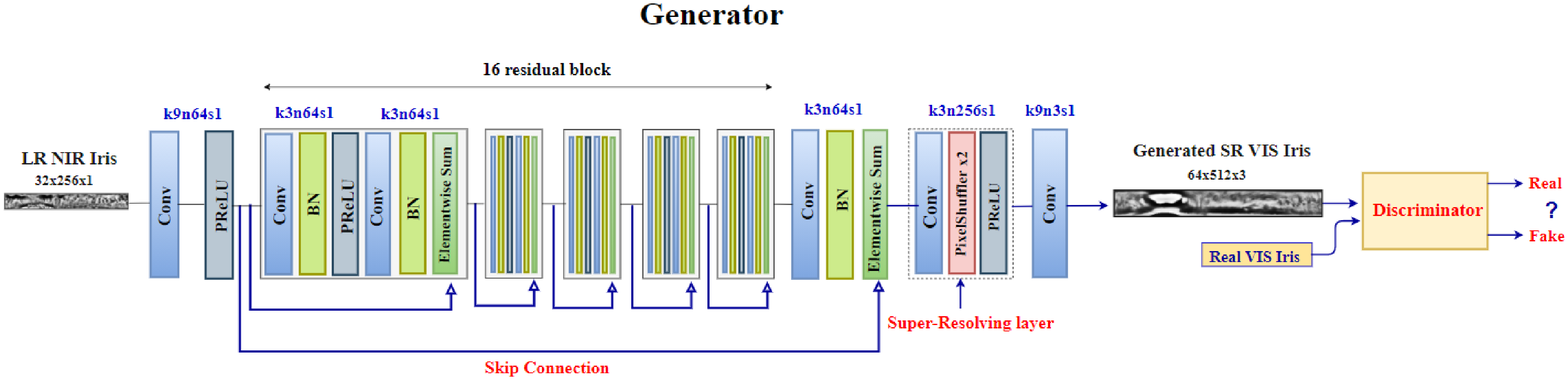}}\\
\subfloat[]{\includegraphics[width=16cm,height=4.0cm]{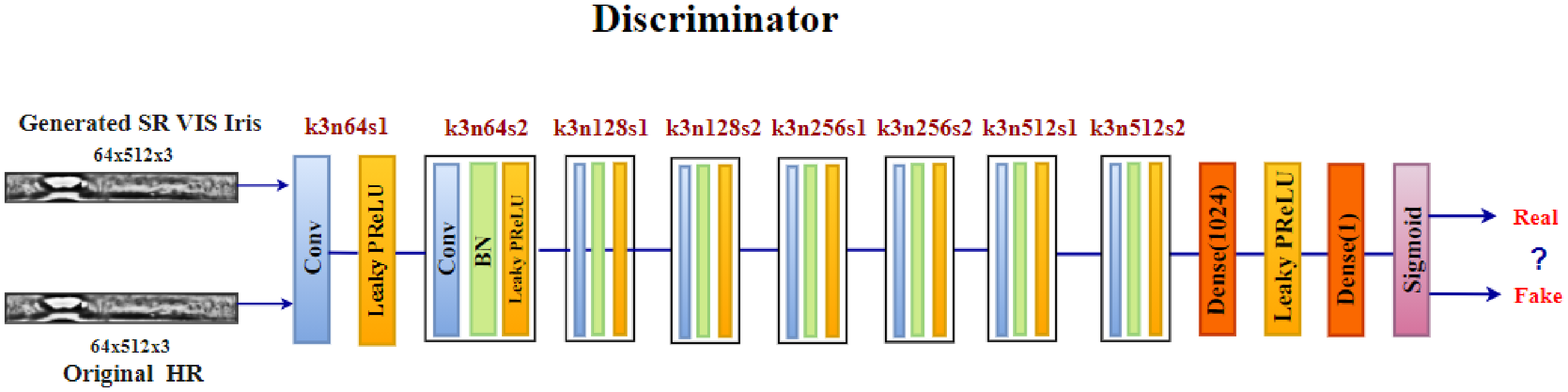}}%
\caption{Architecture of our proposed cGAN (a) generator and (b) discriminator with corresponding kernel size (k), number of feature maps (n) and stride (s) indicated for each convolutional layer.}%
\end{figure*}
\subsubsection{Cross-eyed-cross-spectral iris recognition database}
The Cross-eyed-cross-spectral iris recognition database provides 3,840 iris images from 240 classes for both spectra obtained from 120 subjects. Each of the classes from every subject has eight sample of $400\times300$ pixels for both spectra. We use the same iris segmentation and normalization algorithm used in\textcolor{blue}{\cite{zhao2015accurate}} to normalize all the iris images. The dimension of all the segmented and normalized iris images from this dataset is $512\times64$ pixels. Sample images from the cross-eyed cross-spectral database are shown in \textcolor{blue}{Fig. 3(c)}. In order to ensure fair comparison, we follow the train-test protocol used in\textcolor{blue}{\cite{nalla2016toward}} and choose five image samples for training and the remaining three samples for testing. Applying an all-to-all matching protocol, the network generated 2,160 genuine scores and 516,240 imposter scores. 
\vspace{-0.1cm}
\subsection{cGAN Architecture Implementation}
We adopted our proposed cGAN network structure from reference\textcolor{blue}{\cite{ledig2017photo}} as depicted in \textcolor{blue}{Fig. 4} for our domain translation technique, and formulated the overall loss function inspired
by references\textcolor{blue}{\cite{isola2017image,johnson2016perceptual,ledig2017photo}}. In more detail, for our generator (see \textcolor{blue}{Fig. 4(a)}), we have implemented the ResNet16 architecture\textcolor{blue}{\cite{he2016deep}}, with 16 identical residual blocks. A single residual block is composed of two convolutional layers with $3\times3$ kernels, 64 feature maps, batch-normalization
layers and a Parametric Rectified Linear Unit  (ReLU)\textcolor{blue}{\cite{he2015delving, ioffe2015batch}} activation function. We use this network for iris domain translation based on two different cross-spectral situations (see \textcolor{blue}{Scenario 1} and \textcolor{blue}{Scenario 3} in \textcolor{blue}{Fig. 1}) that we have proposed in our first approach. We also integrate the super-resolution process in the translation network by adding a sub-pixel convolution layer with the layout explained in\textcolor{blue}{\cite{shi2016real}}, which has been illustrated in \textcolor{blue}{Scenario 2(a)} of \textcolor{blue}{Fig. 1}.    
Like\textcolor{blue}{\cite{ledig2017photo}}, for our discriminator architecture, we follow what is presented in\textcolor{blue}{\cite{radford2015unsupervised}}, which consists of eight convolutional
layers with $3\times3$ kernel size. The number of kernels increases from 64 to 512, similar to VGGNet\textcolor{blue}{\cite{simonyan2014very}}. Rather than max-pooling, strided convolution is employed for resolution reduction. As shown in \textcolor{blue}{Fig. 4(b)}, after that, we add a dense layer, a Leaky RELU, another dense layer, and finally, a sigmoid activation function.
In summary, the generator gets a low-resolution (or high-resolution) image from one of the domains and translates it or jointly translates
and super-resolves it to the other domain, and the discriminator is fed with the output of the generator and also a high-resolution
image of the other domain. 


\subsection{cpGAN Architecture Implementation}
We have implemented our cpGAN architecture using the U-Net architecture as the generator module. We have followed the architecture of ResNet-18\textcolor{blue}{\cite{he2016deep}} to implement both the encoder and decoder sections of the U-Net model. In encoder, each block is designed by applying two $3\times3$ convolutions, each followed by a ReLU. For downsampling, it uses a $2\times2$ max pooling operation with stride 2. We double the number of feature channels at each downsampling step. Similarly, each step in the decoder section upscales the feature map by applying a $2\times2$ transpose convolution\textcolor{blue}{\cite{shi2016deconvolution}}, upsampling the dimension of the feature map. Each feature map is concatenated with the corresponding feature map from the encoder, followed by two $3\times3$ convolutions with a ReLU activation function. 
\subsection{Training details}
Both of our frameworks have been implemented in Pytorch. We trained the network with a batch size of 16 and a learning rate of $2\times10^{-4}$. We used the Adam optimizer\textcolor{blue}{\cite{diederik6980method}} with a first-order momentum of 0.5, and a second-order momentum of 0.999. We have used Leaky ReLU as the activation function with a slope of 0.35 for the discriminator. To find the optimal hyper-parameters for our learning algorithms, we have used a random search strategy\textcolor{blue}{\cite{bergstra2012random}}. Following their technique, we experiment with different scaling heuristics to find the optimal hyper-parameter multiplier, which results in the best verification accuracy. Accordingly,
for the network convergence, we set $\lambda_3$ to 1, and $\lambda_4$, and $\lambda_5$ to 0.3. In addition, $\lambda_1$, and $\lambda_2$, are set to $10^{-6}$ and $2\times10^{-3}$, respectively. \\
For training, genuine/impostor pairs are created from the VIS and NIR iris images of the same/different subjects. During the experiments, we ensure that  the training set is balanced by using the same number of genuine and impostor pairs.

\begin{figure*}%
\hspace*{-0.5cm}   
\centering
\subfloat[]{\includegraphics[width=6.5cm,height=6.15cm]{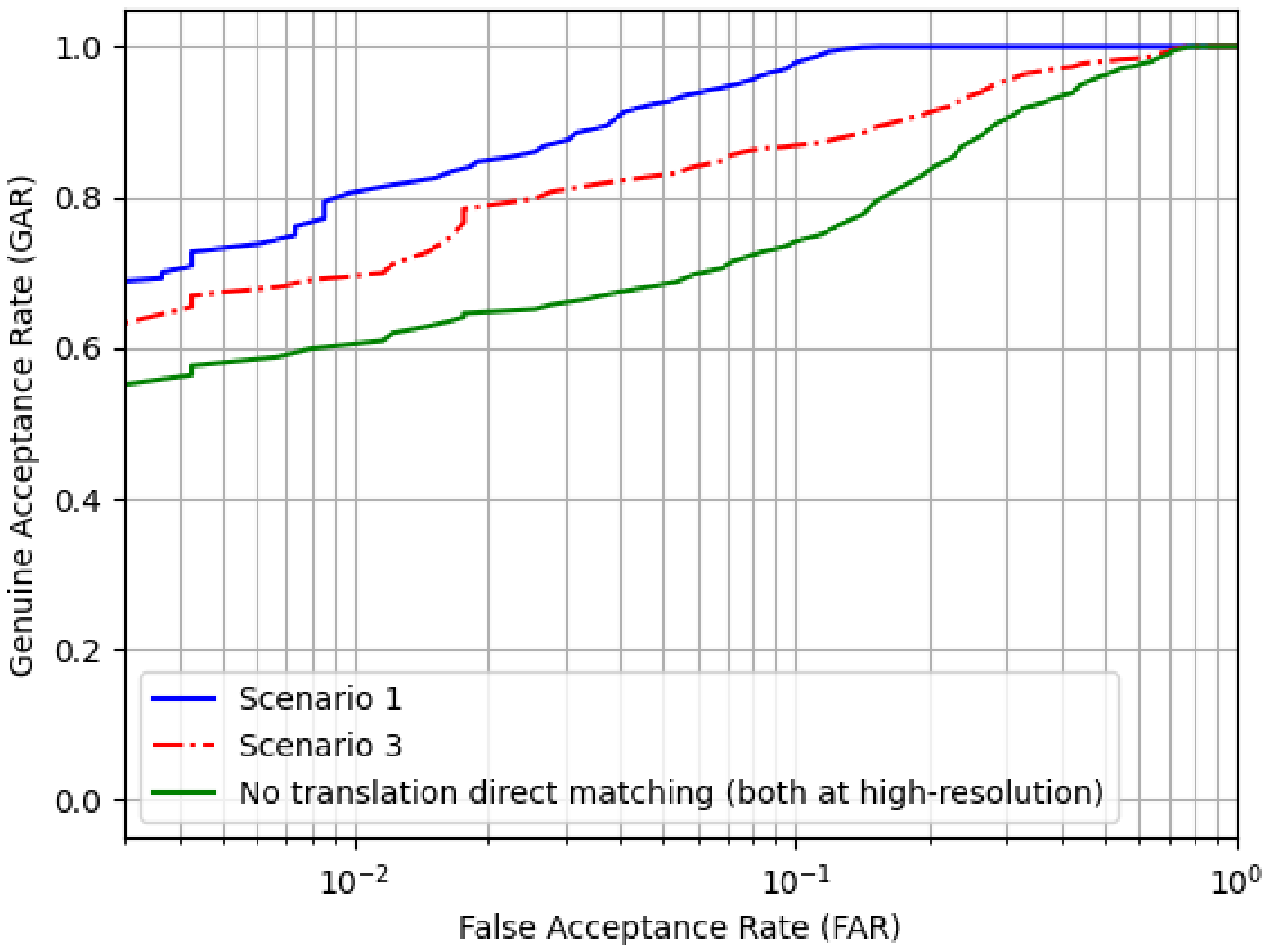}}
\subfloat[]{\includegraphics[width=6.5cm,height=6.15cm]{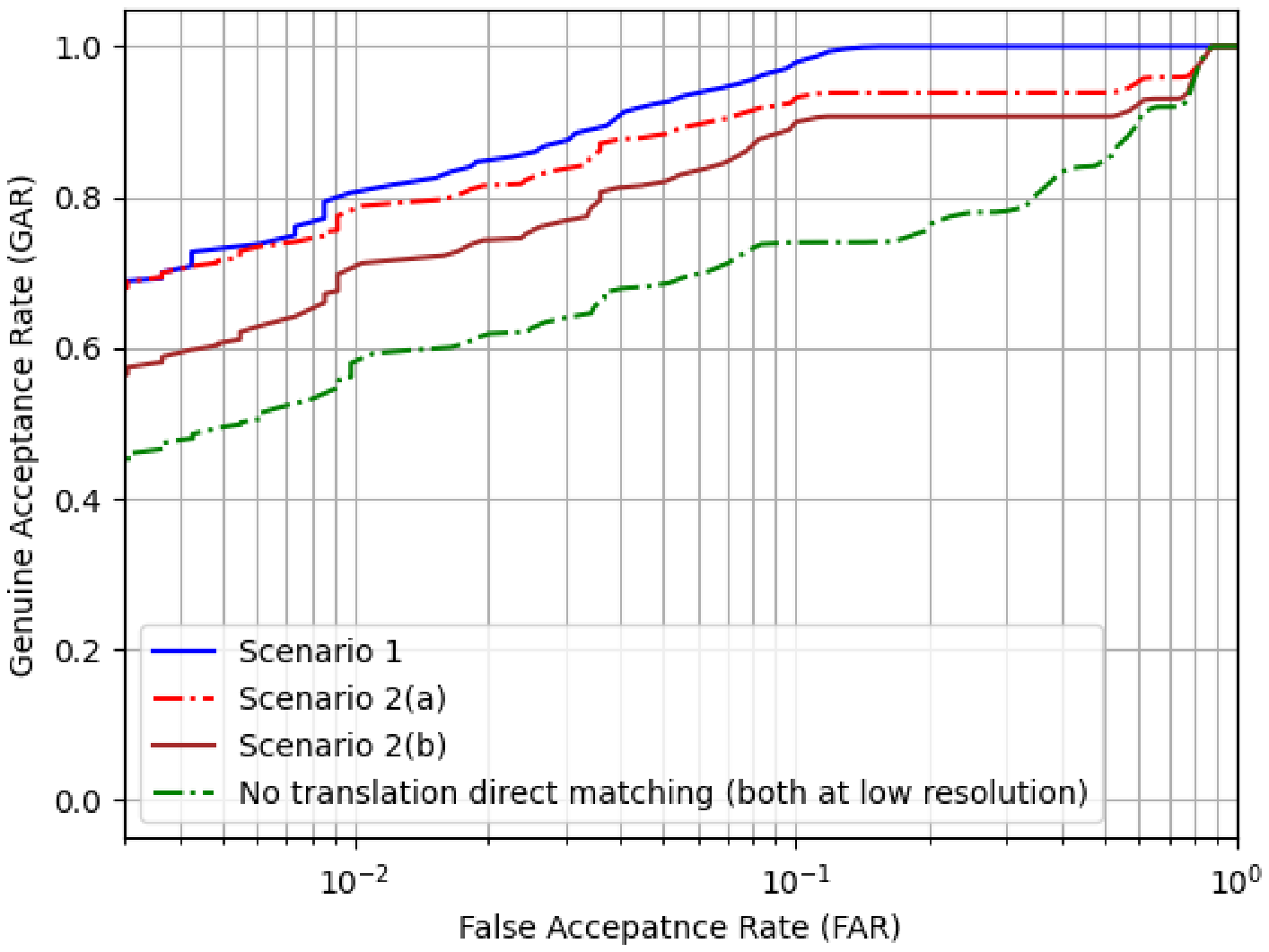}}%
\subfloat[]{\includegraphics[width=6.5cm,height=6.15cm]{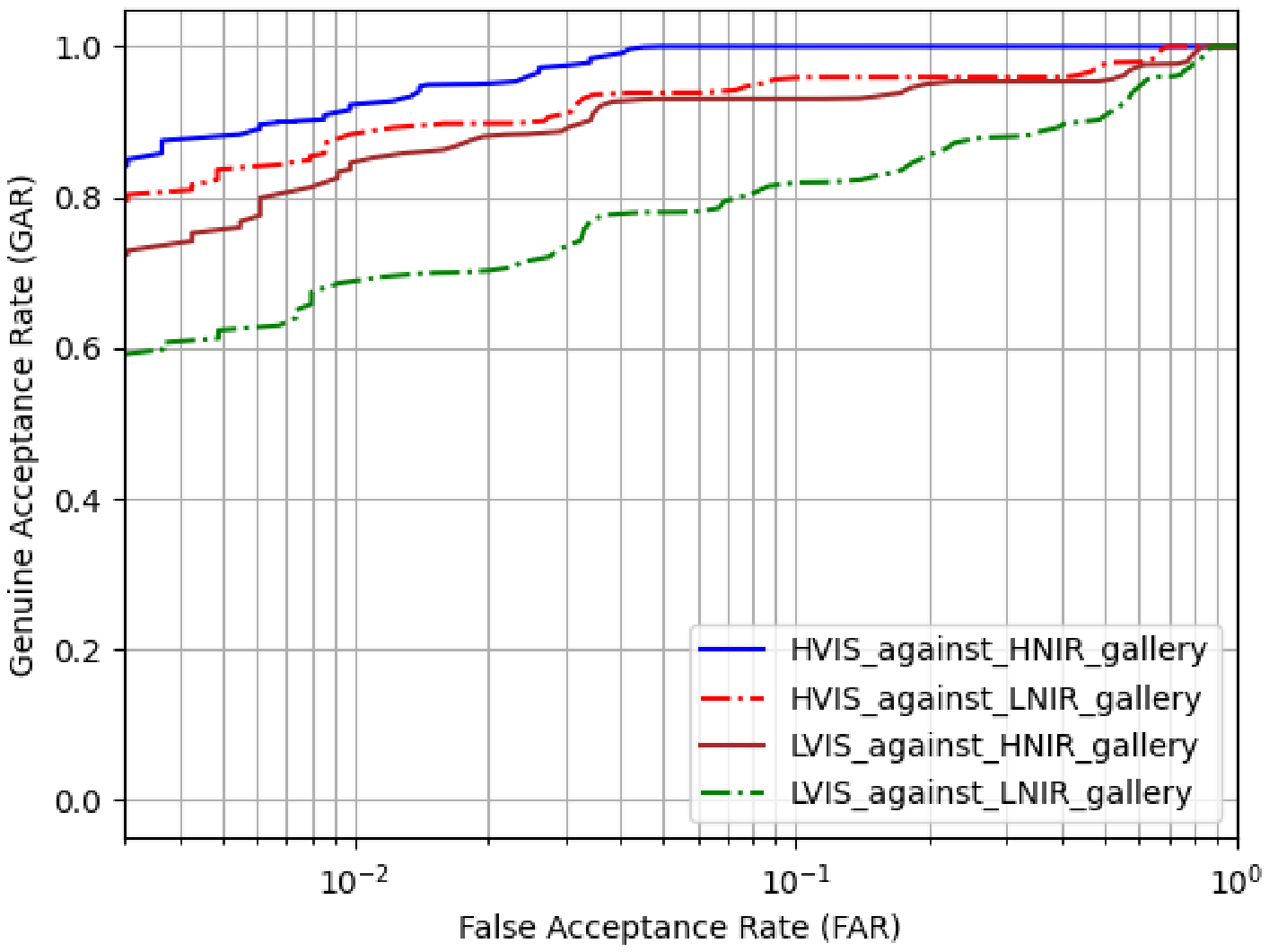}}
\caption{ROC plots showing the performance of our approach (a) Scenario 1 and 3 with cGAN architecture, (b) Scenario 2(a) and 2(b) with cGAN architecture and (c) cpGAN architecture obtained on the PolyU Bi-Spectral database for the different cross-spectral matching scenarios\textcolor{blue}{\cite{mostofa2020cross}}.}%
\end{figure*}

\begin{table*}[t]
\large
\centering
\caption{Comparative performances on the PolyU Bi-Spectral database. Symbol '-' indicates that the metric is not available for that protocol.}
\renewcommand{\arraystretch}{1}
\scalebox{0.60}{\begin{tabular}{c c c c c c}

 \hline
Algorithm&Matching&GAR@FAR=0.1&GAR@FAR=0.01&GAR@FAR=0.001&EER (\%) \\ \hline
Wang et al. \textcolor{blue}{\cite{wang2019toward}}&HR VIS vs HR NIR&|&59.10&37.00&17.03 \\ \hline
CNN with SDH \textcolor{blue}{\cite{wang2019cross}} &HR VIS vs HR NIR&|&90.71&84.50&5.39 \\ \hline
Nalla et al.\textcolor{blue}{\cite{nalla2016toward}}&HR VIS vs HR NIR&58.8&|&|&26.68 \\ \hline
NIR to VIS texture
synthesis using MRF
model \textcolor{blue}{\cite{nalla2016toward}}&HR VIS vs HR NIR&61.91&|&|&23.87 \\ \hline
IrisCode using 1D
Log-Gabor filter\textcolor{blue}{\cite{masek2003matlab}}&HR VIS vs HR NIR&52.6&|&|&17.03 \\ \hline
\textbf{cpGAN}\textcolor{blue}{\cite{mostofa2020cross}} &\textbf{HR VIS vs HR NIR}&\textbf{99.99}&\textbf{92.38}&\textbf{84.98}&\textbf{1.02}\\ \hline
cpGAN\textcolor{blue}{\cite{mostofa2020cross}}&{HR VIS vs LR NIR}&96.5&{89.89}&{81.21}&{1.21} \\ \hline
cpGAN\textcolor{blue}{\cite{mostofa2020cross}}&HR NIR vs LR VIS&93.30&84.75&73.45&1.26 \\ \hline
cpGAN\textcolor{blue}{\cite{mostofa2020cross}}&LR NIR vs LR VIS&82.60&70.10 & 59.97&2.51 \\
\hline
NIR to VIS domain translation (Ours cGAN)&Scenario1&99.50&80.50 & 70.1&1.5\\
\hline
Joint domain translation \& super-resolution (Ours Modified cGAN)&Scenario 2(a)&92.83&75.70 & 69.9&1.6\\
\hline
Domain Translation \& super-resolution (Separate Training)&Scenario 2(b)&88.89&70.10 & 56.10&1.9\\
\hline
VIS to NIR domain translation (Ours cGAN) &Scenario 3&87.49&69.50 & 64.90&1.4\\
\hline
Iriscode (OSIRIS)\textcolor{blue}{\cite{othman2016osiris}} &HR VIS vs HR NIR &74.60&61.10 & 54.50&2.59\\
\hline
Iriscode (OSIRIS)\textcolor{blue}{\cite{othman2016osiris}} &LR VIS vs LR NIR &71.05&55.60 & 43.10&3.0\\
\hline
\end{tabular}}
\end{table*}

\subsection{Evaluation on PolyU Bi-Spectral Database }
We perform our first set of experiments on the PolyU Bi-Spectral database considering many different cross-spectral iris matching cases for both previously-mentioned approaches. In all the experiments, each iris probe image is matched against a gallery of iris images, which generates genuine and imposter scores. Using these matching scenarios, we calculate the key recognition performance parameters, such as genuine acceptance rate (GAR), false acceptance rate (FAR), and equal error rate (EER). In addition, we also plot receiver operating characteristics (ROC) curves to analyze the GAR with respect to FAR. In addition, we compare our results over other considered state-of-the-art cross-spectral iris recognition methods described in\textcolor{blue}{\cite{nalla2016toward,wang2019cross,masek2003matlab}} and\textcolor{blue}{\cite{wang2019toward}} . 
We use the same train-test protocol provided in their original paper for fair comparison. 
\subsubsection{To evaluate the effectiveness of our proposed cGAN architecture, we conduct the following experiments:}
\vspace{0.1cm}
\textbf{(a) Scenario 1 : NIR to VIS domain translation}\\
In this experiment, we train the network to translate a gallery of NIR iris images to its corresponding gallery of synthesized visible iris images at the same resolution (see \textcolor{blue}{Fig. 1 Scenario 1}). Then, each VIS iris probe of the test set is matched against this synthesized VIS iris gallery. We have shown the ROC result from this experiment in \textcolor{blue}{Fig. 5(a)} and report the EER in \textcolor{blue}{Table 2}. 
We observe that our proposed algorithm achieves 99.50\% and 80.50\% GAR at 0.1 and 0.01 FAR, respectively, and obtains an EER of 1.5\%, which outperform the results reported for the algorithms evaluated in\textcolor{blue}{\cite{nalla2016toward,masek2003matlab}}, and\textcolor{blue}{\cite{wang2019toward}} using the same train-test protocol. The network shows significant improvement in cross-spectral iris matching by obtaining 15.53\% and 25.18\% less EER compared to the results in\textcolor{blue}{\cite{wang2019toward}}, and\textcolor{blue}{\cite{nalla2016toward}}, respectively. \\
\textbf{(b) Scenario 2(a) :
Joint translation and super-resolution from the LR NIR to HR VIS domain}
\\
Recently, with the emergence of new biometrics applications on smartphones, there is a strong demand for acquiring high-resolution visible iris images at low cost. However,
while the availability of higher resolution visible iris images will eventually lead to a cross-resolution mismatch in the problem of cross-spectral iris matching, almost no attention has been turned toward it yet. Although there would be higher noise levels in the
visible domain compared to the NIR domain, hopefully the higher resolution can compensate for the effect of this noise. To
address the resolution differences, we determined how to match LR NIR iris images against the HR visible iris
images (i.e., unrolled NIR image size: 32x256, unrolled visible image size: 64x512). We train the network to translate the LR
NIR images to HR VIS
images in such a way that it jointly transforms the image domain
and super-resolves it. Therefore, the network simultaneously learns both image translation and super-resolution tasks. The
network super resolves the input image by a factor of two, and then the output can be used as a gallery of visible iris images for visible-to-visible iris verification. \textcolor{blue}{Fig. 5(b)} and \textcolor{blue}{Table 2} illustrates that our proposed joint translation and super-resolution technique outperforms the baseline approach. It is worth mentioning that we separately train both networks and report the results as a baseline approach to show the comparative performance of the joint learning. We notice that the joint training significantly increases the matching accuracy by 3.94\%, 5.60\% and 13.8\% GAR at FAR of 0.1, 0.01 and 0.001, respectively. \\
\textbf{(c) Scenario 2(b) :
Separate translation and super-resolution from the LR NIR to HR VIS domain}\\ We have also fed the low-resolution NIR images to a cross-domain translation network from reference\textcolor{blue}{\cite{isola2017image}} 
and then the 
low-resolution output is fed to a super-resolution GAN (SRGAN) from reference\textcolor{blue}{\cite{ledig2017photo}}.
This is the \textcolor{blue}{Scenario 2(b)} in \textcolor{blue}{Fig. 1}, and results are shown in \textcolor{blue}{Fig. 5(b)} and \textcolor{blue}{Table 2}. The separate training achieves 88.89\%, 70.10\%, and 56.10\% GAR at FAR of 0.1, 0.01 and 0.001, respectively, which are significantly lower compared to the joint training. These results validate our idea of joint transformation and super-resolution. \\
\textbf{(d) Scenario 3 :
VIS to NIR Domain Transformation}
\\
In order to examine whether or not the NIR-to-visible image translation is a more effective solution than translating the
visible to NIR, both at the same resolution, we have trained a network to map the visible images to the NIR domain and
performed verification on the synthesized NIR iris images (i.e., matching the synthesized NIR images against a gallery of
NIR images). We feed a given visible iris probe image to the network, which is trained to map visible to NIR images, and
then use the output image to compare with an existing gallery of NIR images. We report the ROC result obtained from this experiment in \textcolor{blue}{Fig. 5(b)} along with the comparative results from other approaches. We consider the algorithm used in\textcolor{blue}{\cite{othman2016osiris}} as comparable benchmark for this scenario. It proves the efficacy of our proposed approach by acquiring 2.19\% less EER compared to the baseline result mentioned above. 
\subsubsection{Similarly, to ascertain true cross-spectral matching ability of our proposed cpGAN network, we experiment with different types of cross-comparisons as follows:}
\textbf{(a) Matching HR VIS probe against a HR NIR gallery}:\\
To perform this verification, we train our coupled learning network with the unrolled HR $64\times512$ VIS and NIR iris images such that  VIS and NIR generators are trained to
obtain domain invariant features in a common latent embedding
subspace using a contrastive loss. We plot ROC curves comparing our approach with other state-of-the-art deep learning methods presented in
\textcolor{blue}{\cite{wang2019toward,wang2019cross}}, which apply different types of feature extraction techniques. From \textcolor{blue}{Fig. 5 (c)} and \textcolor{blue}{Table 2}, we notice that our cpGAN framework performs much better than the baseline matching algorithms mentioned above. In this setting, our method achieves 1.67\%
more identification accuracy with 4.37\% decrease in EER
compared to the most recent cross-spectral iris recognition
method\textcolor{blue}{\cite{wang2019cross}}. Additionally, it outperforms the method described in\textcolor{blue}{\cite{othman2016osiris,nalla2016toward}} by a significant decrease of 1.57\% and
22.85\% in EER, respectively. This significant improvement clearly indicates that using a cpGAN framework for projecting both the VIS and NIR iris images into a common latent embedding subspace
to retrieve the domain invariant features
is better than the other existing deep learning methods.\\
\textbf{(b) Matching HR VIS probe against a LR NIR gallery:}\\
Here, we consider a realistic iris matching scenario to analyze the cross-spectral matching accuracy of our network. Due to the advances in imaging technology, opportunistic iris images extracted from faces in the visible spectrum are at a higher resolution, while images already stored in the gallery are in the low-resolution NIR domain. It has become a challenging task to build a correlation between iris images in different resolutions as well as in different spectra. Many algorithms fail to retrieve accurate semantic similarity among iris images of different resolutions and spectra, which has resulted in a significant performance degradation in existing iris verification systems.
Therefore, we resolve this issue by training our cpGAN with the unrolled HR ($64\times512$)
VIS and LR ($32\times256$) NIR iris images, which ensures the retrieval of contextual and semantic features of the iris images in a common embedding subspace. The results summarized in \textcolor{blue}{Fig. 5(c)} and \textcolor{blue}{Table 2} indicate that the cpGAN network remains robust enough to provide superior results compared to our matching \textcolor{blue}{Scenario 2(a)} that was shown in Fig. 5(b). It has increased the GAR almost by 14\% at 0.01 FAR.\vspace{0.5cm}\\
\textbf{(c) Matching LR VIS iris images against a gallery of HR NIR iris images:}\\
In addition to the study mentioned above, we have also focused on matching LR VIS iris probe against a gallery of HR NIR iris images. We consider a fact when subjects are at a large standoff distance from the camera. Consequently, captured faces are assumed to be suffering from poor quality due to low-resolution.
On the other hand, the gallery images have comparatively higher resolution which are usually taken in the NIR spectrum. Therefore, the modality gap between probe and gallery images makes the cross-spectral matching even more challenging. Hence, we train the VIS and NIR generator of our network with the unrolled LR VIS iris images ($32\times256$) and HR NIR iris images $(64\times512$), respectively, and perform matching in the latent embedded subspace, that contains basic information about the iris texture patterns irrespective of the resolution. The experimental results reported in \textcolor{blue}{Table 2} show that our proposed scheme has produced EER with a value of 1.26\% which proves the adequacy of our approach even in low-quality videos.\\ 

\noindent \textbf{(d) Matching LR VIS iris images against a
gallery of LR NIR iris images :}\\
We also perform additional experiments where our gallery images are in the low-resolution NIR domain. To investigate the matching performance of our network, we feed both the VIS and NIR generator with the unrolled LR VIS and NIR iris images. 
The experimental results reported in \textcolor{blue}{Table 2} and \textcolor{blue}{Fig. 5(c)} indicate the matching accuracy of our network for this cross-spectral setting compared to the approach used in\textcolor{blue}{\cite{othman2016osiris}}. Even though we achieve an EER of 2.51\% that is much lower than several comparable methods, there is a tradeoff with verification performance, which is not as satisfactory as our previous experiments outlined above.

\begin{figure*}%
\hspace*{-0.5cm}   
\centering
\subfloat[]{\includegraphics[width=6.5cm,height=6.15cm]{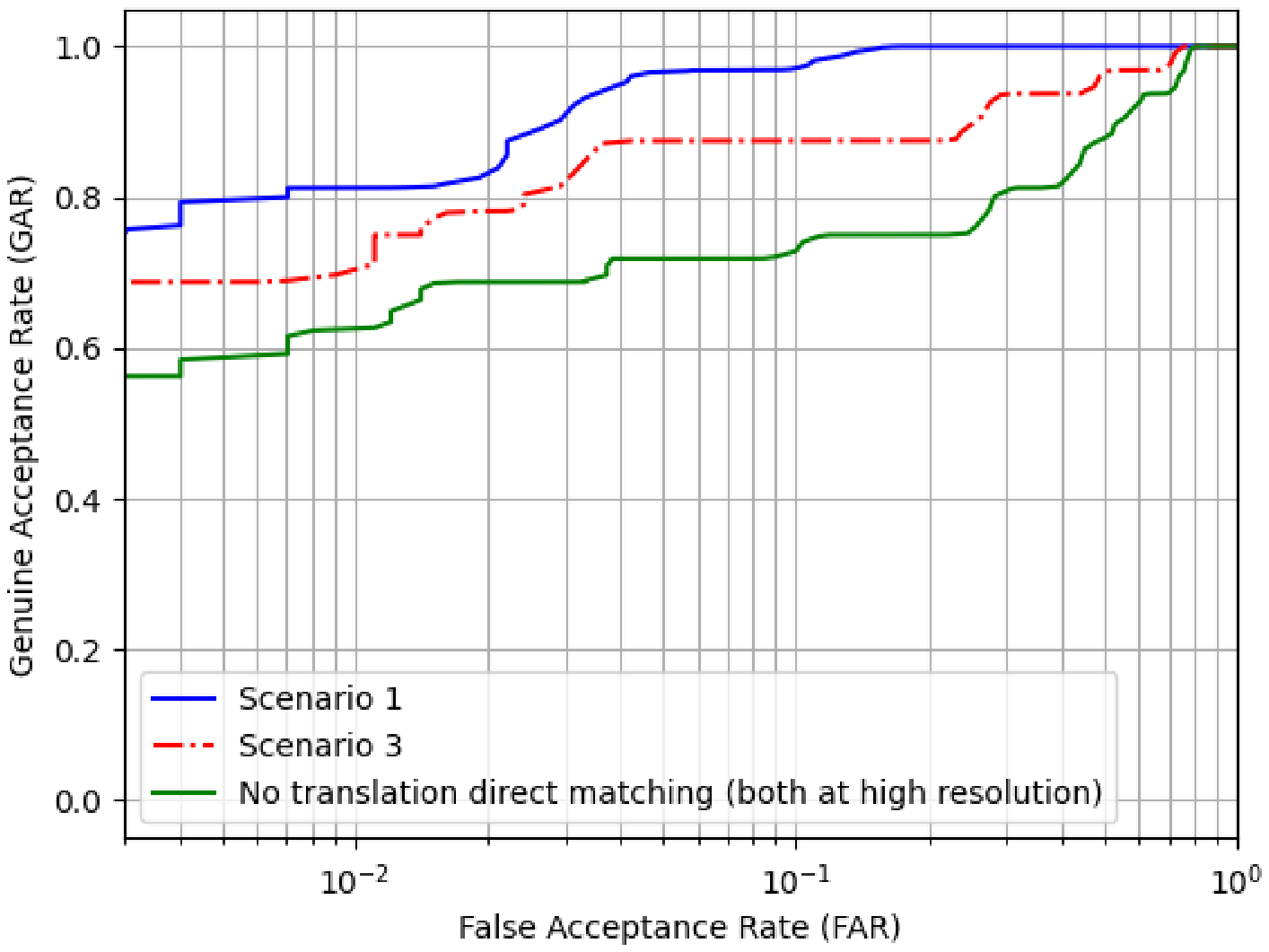}}
\subfloat[]{\includegraphics[width=6.5cm,height=6.15cm]{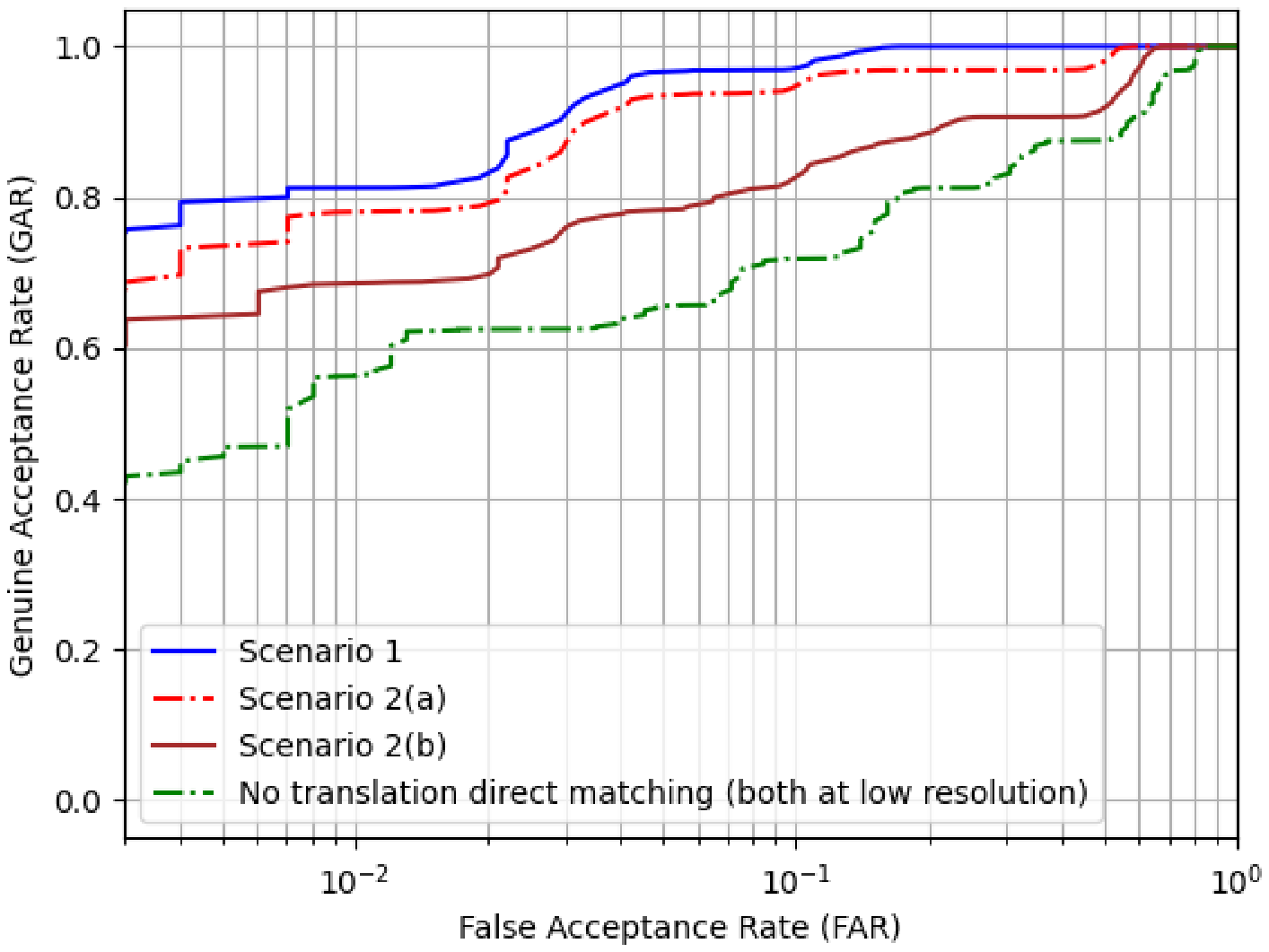}}%
\subfloat[]{\includegraphics[width=6.5cm,height=6.15cm]{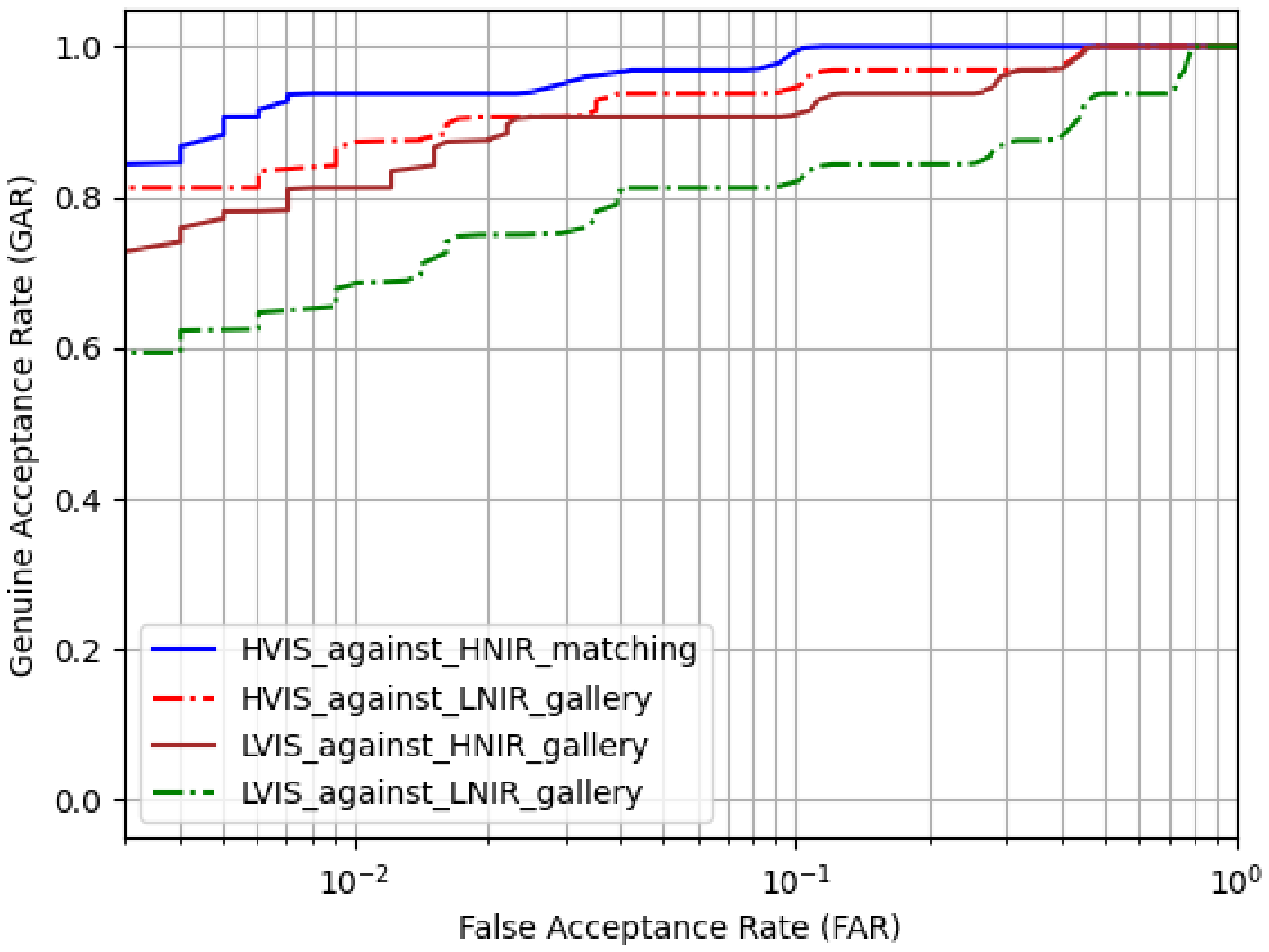}}
\caption{ROC plots showing the performance of our approach (a) Scenario 1 and 3 with cGAN architecture, (b) Scenario 2(a) and 2(b) with cGAN architecture and (c) cpGAN architecture obtained on the WVU face and iris database for the different cross-spectral matching scenarios.}%
\end{figure*}

\begin{table*}[t]
\large
\centering
\caption{Comparative performances on the WVU face and iris database.}
\renewcommand{\arraystretch}{1}
\scalebox{0.60}{\begin{tabular}{c c c c c c}

 \hline
Algorithm&Matching&GAR@FAR=0.1&GAR@FAR=0.01&GAR@FAR=0.001&EER (\%) \\ \hline
\textbf{cpGAN(Ours)} &\textbf{HR VIS vs HR NIR}&\textbf{99.54}&\textbf{93}&\textbf{84}&\textbf{0.90}\\ \hline
{cpGAN(Ours)}&{HR VIS vs LR NIR}&97.04&{87.7}&{80.8}&{1.15} \\ \hline
cpGAN(Ours)&HR NIR vs LR VIS&92.89&83.50&72.70&1.20 \\ \hline
cpGAN(Ours)&LR NIR vs LR VIS&82.52&69.2 & 59.70&1.85 \\
\hline
NIR to VIS domain translation (Ours cGAN)&Scenario 1&97.79&80.8 & 75.1&1.0\\
\hline
Joint domain translation \& super-resolution (Ours Modified cGAN)&Scenario 2(a)&94.97&77.8 & 69.5&1.34\\
\hline
Domain translation \& super-resolution (Separate Training)&Scenario 2(b)&83.50&69.60 & 60.0&1.97\\
\hline
VIS to NIR domain translation (Ours cGAN) &Scenario 3&88.53&70.10 & 67.70&1.38\\
\hline
Iriscode (OSIRIS)\textcolor{blue}{\cite{othman2016osiris}} &HR VIS vs HR NIR &76.02&62.0 & 56.1&2.14\\
\hline
Iriscode (OSIRIS)\textcolor{blue}{\cite{othman2016osiris}} &LR VIS vs LR NIR &71.7&55.5 & 42.7&3.01\\
\hline
\end{tabular}}
\end{table*}
\subsection{Evaluation on WVU Face and Iris Database}
To assess the effectiveness of our proposed approaches, we conduct a number of extensive experiments on the WVU face and iris database for different cross-spectral matching scenarios similar to the experiments performed on the PolyU bi-spectral database. To the best of our knowledge, there is no other baseline algorithm in the literature that have performed cross-spectral iris matching on this dataset.
Therefore, our evaluation on the WVU face and iris dataset yields a new state-of-the-art cross-spectral iris matching result, which will further encourage the biometric research community to investigate the performance of other existing algorithms on this dataset. In this context, we first report on the evaluation of the method in our first approach for matching cross-spectral iris images under the same spectral domain. Then we discuss experimental results obtained from our second method, which performs matching in the embedded domain.

\begin{figure*}%
\hspace*{-0.5cm}   
\centering
\subfloat[]{\includegraphics[width=6.5cm,height=6.15cm]{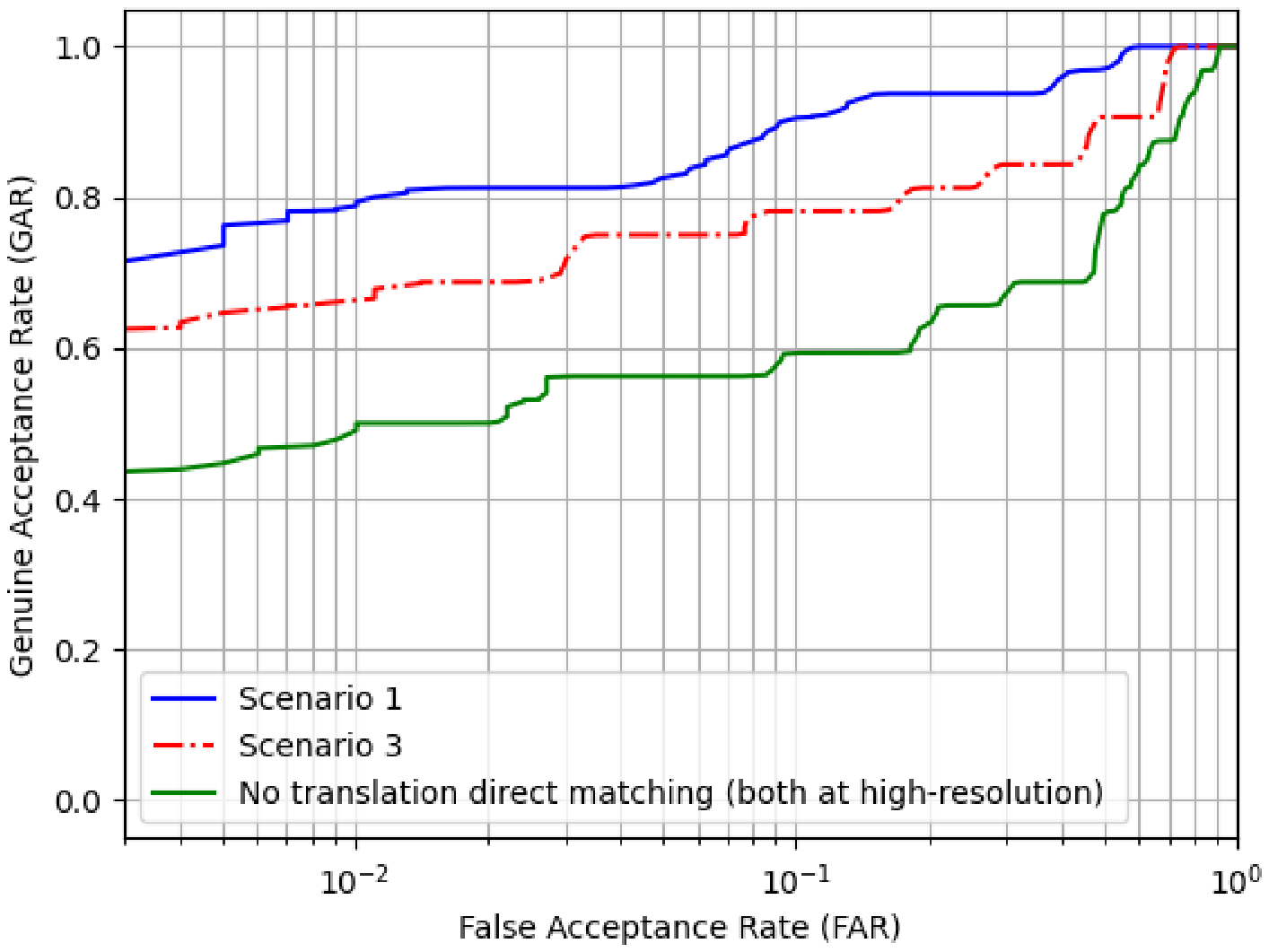}}
\subfloat[]{\includegraphics[width=6.5cm,height=6.15cm]{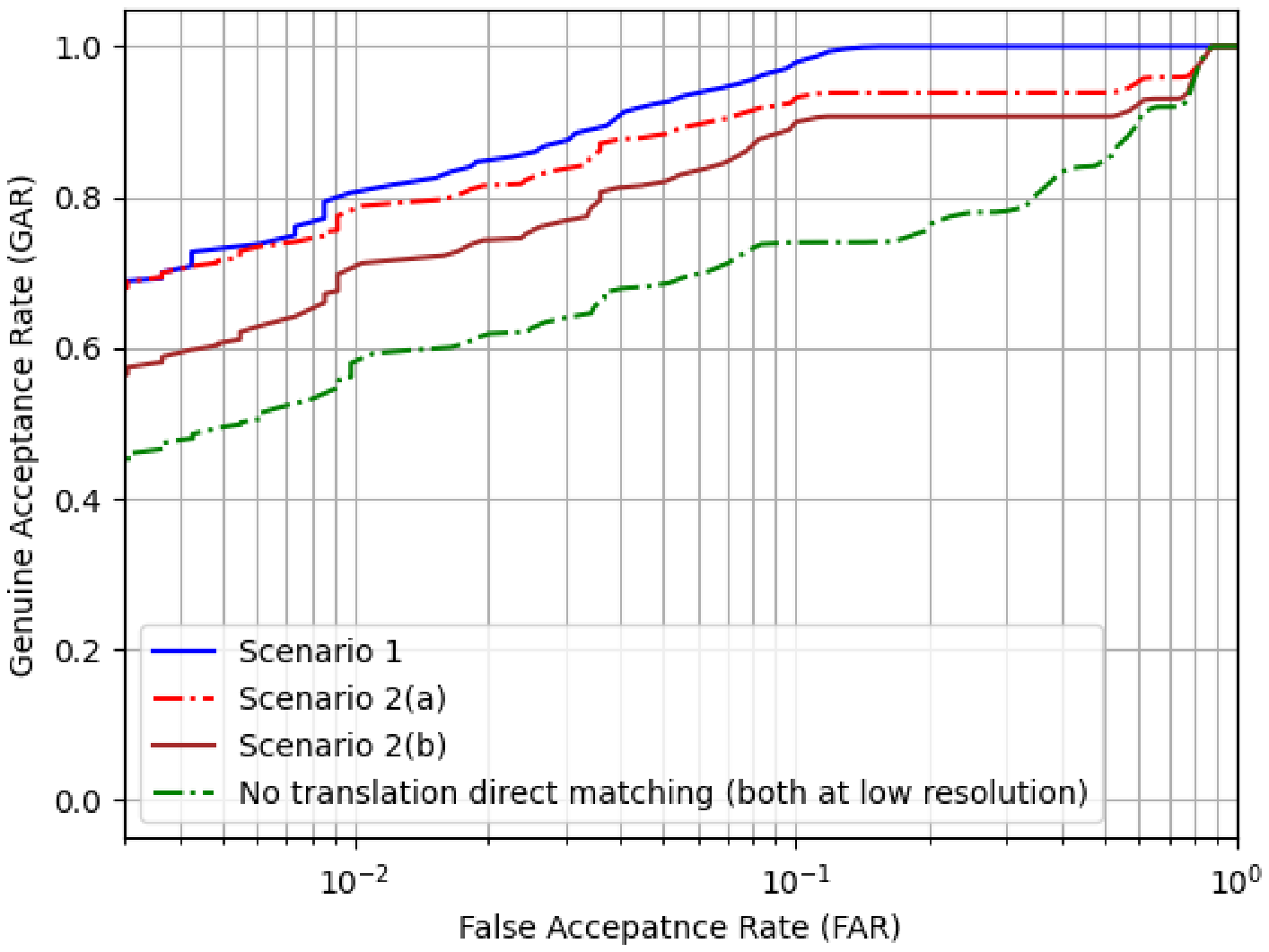}}%
\subfloat[]{\includegraphics[width=6.5cm,height=6.15cm]{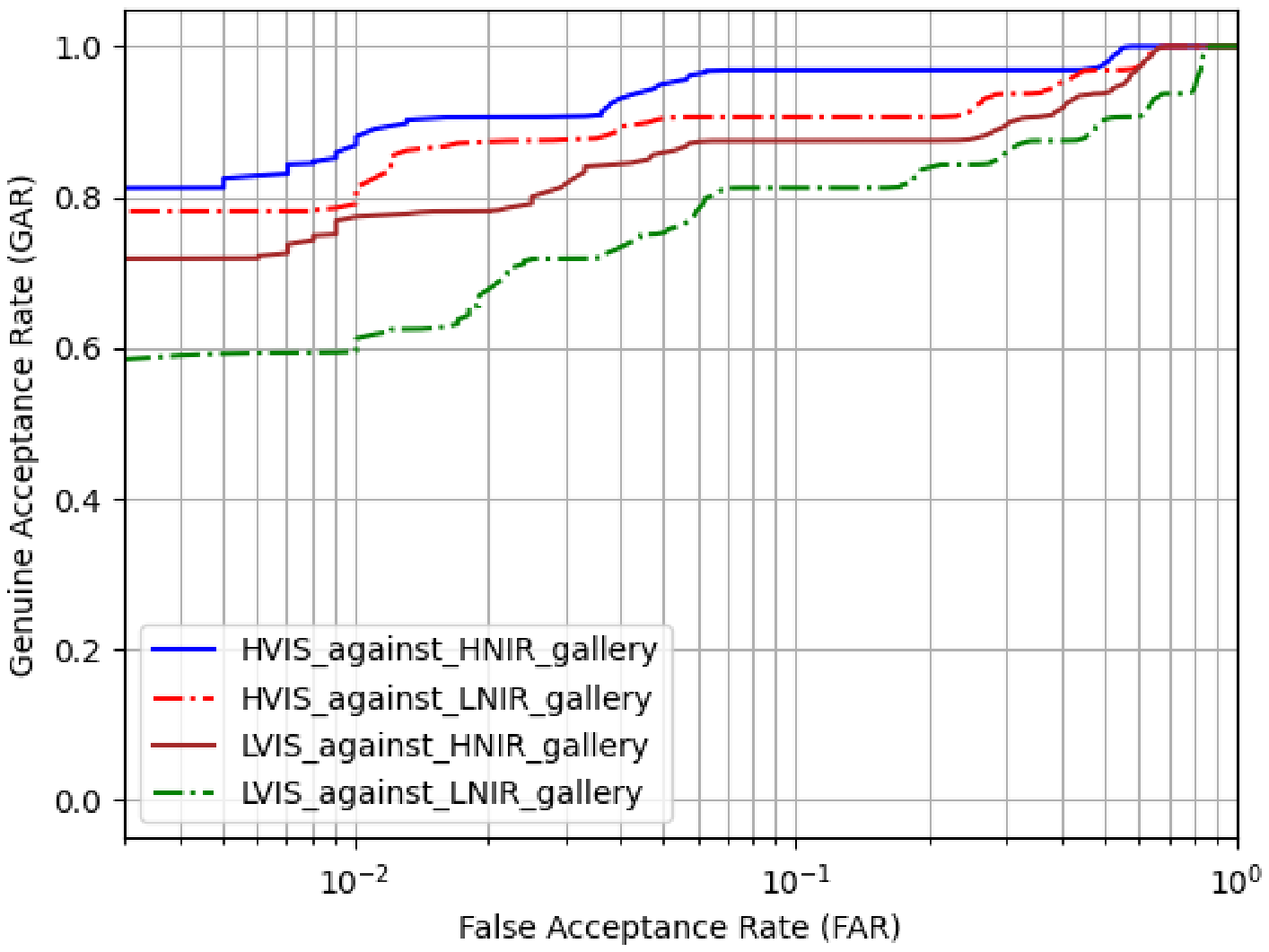}}
\caption{ROC plots showing the performance of our approach (a) Scenario 1 and 3 with cGAN architecture, (b) Scenario 2(a) and 2(b) with cGAN architecture and (c) cpGAN architecture obtained on the Cross-eyed-cross-spectral iris recognition database for the different cross-spectral matching scenarios.}%
\end{figure*}

\subsubsection{Matching Results Obtained From cGAN Architecture}

We consider similar experimental scenarios as stated in Section 5.5 when we trained our cGAN model with the WVU face and iris database. We plot ROC results in \textcolor{blue}{Fig. 6(a)} obtained from the cGAN network that has been trained and tested for Scenario 1 and Scenario 3. In addition, we summarize the EER in \textcolor{blue}{Table 3}. \textcolor{blue}{Fig. 6(a)} and \textcolor{blue}{Table 3} demonstrate that our proposed algorithm achieves  99.50\%, 80.50\% and 70.1\% GAR at 0.1, 0.01 and 0.001 FAR, respectively for Scenario 1, where each VIS iris probe image of the test set is matched against a gallery of synthesized VIS iris images. For comparison, we report recognition accuracy for this database which has been obtained from the algorithm used in\textcolor{blue}{\cite{othman2016osiris}} for matching the HR VIS iris probe image against a gallery of HR NIR iris images. It is obvious that our proposed cGAN algorithm significantly increases the recognition accuracy by 18.8\% for the FAR of 0.01 with 1.14\% decrease in EER compared to the cross-spectral iris matching result reported as a baseline approach (Matching HR VIS iris probe image against a gallery of NIR iris images).

We also report identification accuracy for the reverse case as described by \textcolor{blue}{Scenario 3}. In this case, we train a network to map the VIS iris images to the NIR domain and perform matching between the synthesized NIR iris images and a gallery of NIR iris images. The experimental results summarized in \textcolor{blue}{Table 3} prove that even for the reverse scenario our approach achieves 8.1\% higher recognition accuracy and 0.76\% lower EER compared to the baseline result.

Again, to ascertain the comparative performance of our joint network described in \textcolor{blue}{Scenario 2(a)}, which simultaneously translates and super-resolves a LR NIR iris image to a HR VIS image, we separately train both networks and use the result as baseline. Additionally, we apply the popular IrisCode approach\textcolor{blue}{\cite{othman2016osiris}} to generate comparative matching scores (i.e., matching the LR VIS iris probe against a gallery of LR NIR iris images). The ROC results from this set of experiments are shown in \textcolor{blue}{Fig. 6(b)}, which indicate the superiority of our proposed joint learning method over other benchmark results. \textcolor{blue}{Table 3} summarizes that our joint translation and super-resolution technique significantly outperforms the result obtained from separate training by 8.20\% recognition rate at 0.01 FAR. 

\subsubsection{Matching Results Obtained From cpGAN Architecture}

To evaluate the verification performance of our coupled learning framework, we follow similar experimental settings that were previously discussed in the earlier section for the PolyU bi-spectral database. We experiment with four different cross-spectral and cross-resolution iris matching scenarios for this dataset and plot ROC results in \textcolor{blue}{Fig. 6(c)} to show the recognition accuracy of our proposed network. We also provide EER results in \textcolor{blue}{Table 3}. 

The experimental results illustrated in \textcolor{blue}{Table 3} indicate that our cpGAN network, which performs verification in the embedding subspace, achieves a lower EER of 0.90\% with a higher GAR of 93\% at 0.01 FAR, when matching HR VIS iris probe image against a gallery of HR NIR iris images. Moreover, it significantly improves the matching accuracy by 31\% GAR at 0.01 FAR compared to the reported
baseline result\textcolor{blue}{\cite{othman2016osiris}} using the same test data for the same cross-spectral matching scenario.

Next, we consider a real-life cross-resolution matching scenario within the cross-spectral domain and train our cpGAN with the unrolled HR ($64\times512$) VIS and LR ($32\times256$) NIR iris images, which gradually learns the inherent hidden correlation between iris images in the  cross-resolution and cross-spectral domains. The matching results briefly presented in \textcolor{blue}{Fig. 6(c)} and \textcolor{blue}{Table 3} show that our cpGAN network ensures an accurate retrieval by outperforming the matching \textcolor{blue}{Scenario 2(a)} in \textcolor{blue}{Fig. 6(b)} with 10.9\% higher recognition accuracy at 0.01 FAR.

Also, we conduct experiments for the scenario with low-quality videos. ROC results and EER scores
detailed in \textcolor{blue}{Table 3} prove
that our proposed scheme maintains robust performance even when matching a LR VIS iris probe against an available HR NIR gallery. It has generated an EER of 1.20\%, which is considered as a lower EER value for an ideal biometric system. 

Finally, we investigate the verification performance of our proposed cpGAN network when iris images in the gallery are in low-resolution NIR domain. Therefore, we force the cpGAN network to learn invariant features in the common embedding subspace from both the LR ($32\times256$) VIS and NIR iris images. The experimental results in \textcolor{blue}{Table 3} show that our proposed algorithm obtains 3.7\% more recognition accuracy at 0.01 FAR than the approach used in\textcolor{blue}{\cite{othman2016osiris}} on the same test data for this cross-spectral setting.

\begin{table*}[t]
\large
\centering
\caption{Comparative performances on the Cross-eyed-cross-spectral iris recognition database. Symbol '-' indicates that the metric is not available for that protocol.}
\renewcommand{\arraystretch}{1}
\scalebox{0.60}{\begin{tabular}{c c c c c c}
 \hline
Algorithm&Matching&GAR@FAR=0.1&GAR@FAR=0.01&GAR@FAR=0.001&EER (\%) \\ \hline
CNN with SDH \textcolor{blue}{\cite{wang2019cross}} &HR VIS vs HR NIR&|&87.18&|&6.34 \\ \hline
NIR to VIS texture
synthesis using MRF
model \textcolor{blue}{\cite{nalla2016toward}}&HR VIS vs HR NIR&78.13&|&|&18.40 \\ \hline
IrisCode using 1D
Log-Gabor filter\textcolor{blue}{\cite{masek2003matlab}}&HR VIS vs HR NIR&70.3&|&|&19.48 \\ \hline
\textbf{cpGAN(Ours)} &\textbf{HR VIS vs HR NIR}&\textbf{96.30}&\textbf{89.4}&\textbf{81.8}&\textbf{1.1}\\ \hline
{cpGAN(Ours)}&{HR VIS vs LR NIR}&90.3&{81.7}&{79.6}&{1.28} \\ \hline
cpGAN(Ours)&HR NIR vs LR VIS&86.40&78.4&72.3&1.31 \\ \hline
cpGAN(Ours)&LR NIR vs LR VIS&81.80&62.0 & 59.0&2.55 \\
\hline
NIR to VIS domain translation (Ours cGAN)&Scenario 1&90.30&80.09 & 70.1&1.54\\
\hline
Joint domain translation \& super-resolution (Ours Modified cGAN)&Scenario 2(a)&80.8&74.8 & 67.02&1.71\\
\hline
Domain translation \& super-resolution (Separate Training)&Scenario 2(b)&71.30&60.0 & 54.90&3.04\\
\hline
VIS to NIR domain translation (Ours cGAN) &Scenario 3&79.0&66.8 & 63.8&2.17\\
\hline
Iriscode (OSIRIS)\textcolor{blue}{\cite{othman2016osiris}} &HR VIS vs HR NIR &60.0&51.5 & 44.8&3.9\\
\hline
Iriscode (OSIRIS)\textcolor{blue}{\cite{othman2016osiris}} &LR VIS vs LR NIR &53.1&44.2 & 38.8&5.67\\
\hline
\end{tabular}}
\end{table*}

\subsection{Evaluation on Cross-Eyed-Cross-Spectral Iris Recognition Database}
We perform another set of experiments using the cross-eyed database to quantify the cross-spectral iris recognition accuracy for both of the approaches developed for this paper. We follow the same experimental settings conducted for the other two datasets for different cross-spectral matching scenarios that have been described in the previous sections. It is worth noting that while comparing our results obtained for this dataset over existing algorithms\textcolor{blue}{\cite{nalla2016toward,wang2019cross,masek2003matlab,othman2016osiris}}, we follow the same train-test protocol used in their paper to show fair evaluation. 

The comparative matching results from our cGAN and cpGAN architectures  
are shown in \textcolor{blue}{Fig. 7}, while the corresponding EER results are summarized in \textcolor{blue}{Table 4}. For comparison we use several highly competitive benchmark MRF approach\textcolor{blue}{\cite{nalla2016toward}}, polpular gabor filter based IrisCode\textcolor{blue}{\cite{othman2016osiris}}, SDH method\textcolor{blue}{\cite{wang2019cross}} and another 1D log-gabor filter based IrisCode\textcolor{blue}{\cite{masek2003matlab}} to ascertain the superiority of our proposed approaches. 

\textcolor{blue}{Fig. 7(a)} depicts the experimental results for \textcolor{blue}{Scenario 1} and \textcolor{blue}{Scenario 3} from our cGAN architecture compared to the baseline result using the most widely deployed IrisCode\textcolor{blue}{\cite{othman2016osiris}} approach. The results from \textcolor{blue}{Scenario 1} indicate that our proposed domain translation technique using the cGAN architecture significantly improves the cross-spectral iris matching accuracy by 28.59\% at 0.01 FAR compared to the benchmark result using the IrisCode\textcolor{blue}{\cite{othman2016osiris}} approach. In addition, it also achieves 15.3\% higher GAR at 0.01 FAR and 0.73\% lower EER even when we experiment matching for \textcolor{blue}{Scenario 3}.

In \textcolor{blue}{Fig. 7(b)}, we present ROC results for showing the performance of our proposed joint network \textcolor{blue}{Scenario 2(a)} where the network learns to translate and super-resolve simultaneously from the LR NIR to HR VIS iris image, and compare this result to the approach when both techniques are applied separately (see \textcolor{blue}{Fig. 1 Scenario 2(b)}). \textcolor{blue}{Table 4} shows that joint training obtains 74.8\% GAR at 0.01 FAR, which outperforms the separate training considered as baseline by 14.8\% GAR.\vspace{0.2cm}\\ 
\indent We also investigate the performance of our coupled learning framework for four different cross-spectral and cross-resolution scenarios. We plot the resulting ROC curves in \textcolor{blue}{Fig. 7(c)}. \textcolor{blue}{Table 4} summarizes the EER results comparing our proposed approach with other state-of-the-art deep learning iris recognition method proposed in\textcolor{blue}{\cite{nalla2016toward,masek2003matlab,wang2019cross,othman2016osiris}} for the same train-test protocol. We notice that when we match the HR VIS iris probe image against a HR NIR iris gallery, our cpGAN achieves superior recognition performance over the other baseline matching algorithms. It obtains almost 26\% and 18.17\% more identification accuracy compared to the approach used in\textcolor{blue}{\cite{masek2003matlab}} and \textcolor{blue}{\cite{nalla2016toward}}, respectively.
In addition, it also outperforms the most competitive cross-spectral iris recognition approach\textcolor{blue}{\cite{wang2019cross}} in the literature by a remarkable decrease of 5.24\% in EER. 
All the other scenarios achieve EER less than 2\%, which reveals the robustness of our coupled network. Again, even if we consider a LR NIR iris probe matched against a LR NIR iris gallery, we observe it performs much better than the benchmark using IrisCode\textcolor{blue}{\cite{othman2016osiris}} for the same scenario. 
\vspace{-0.5cm}

\begin{figure*}%
\centering
\subfloat[]{\includegraphics[width=7.5cm,height=5cm]{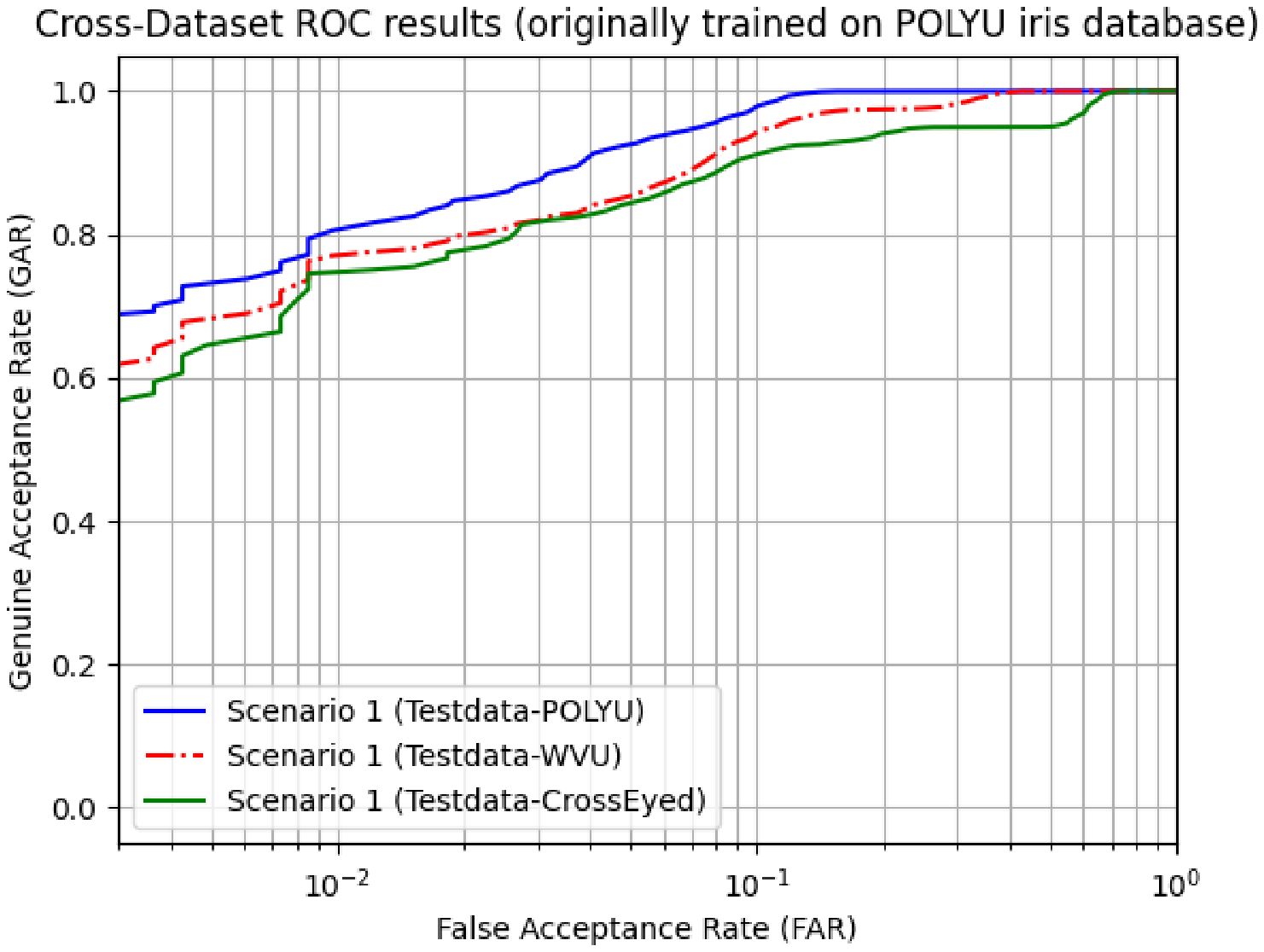}}
\subfloat[]{\includegraphics[width=7.5cm,height=5cm]{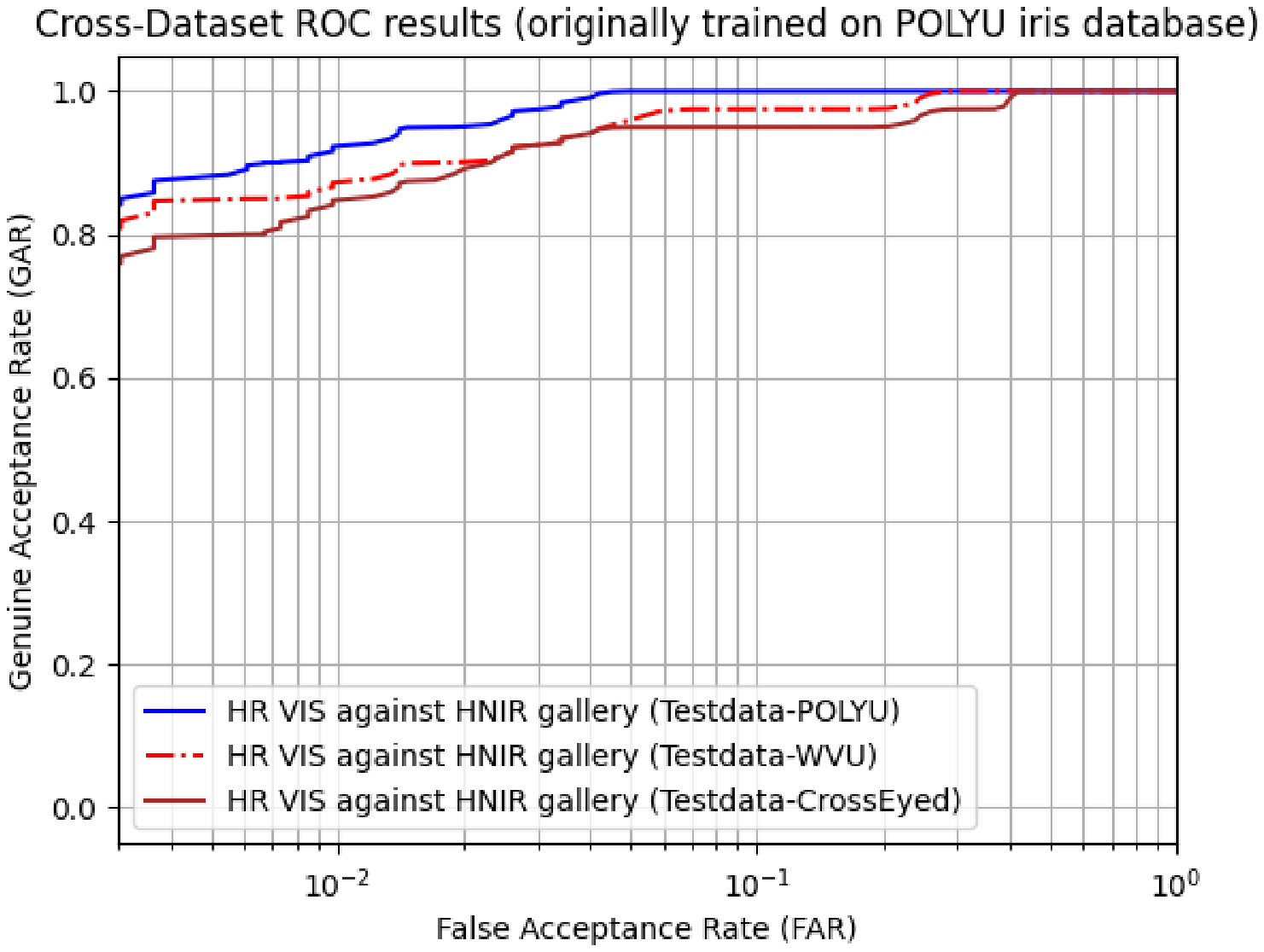}}
\caption{Comparative ROC results showing the cross-database matching of our approach (a) Scenario 1 with cGAN architecture (b) cpGAN architecture (matching the HR VIS iris probe against a HR NIR gallery) where both networks were trained only on the PolyU bi-spectral dataset.}%
\end{figure*}

\begin{figure*}%
\centering
\subfloat[]{\includegraphics[width=7.5cm,height=5cm]{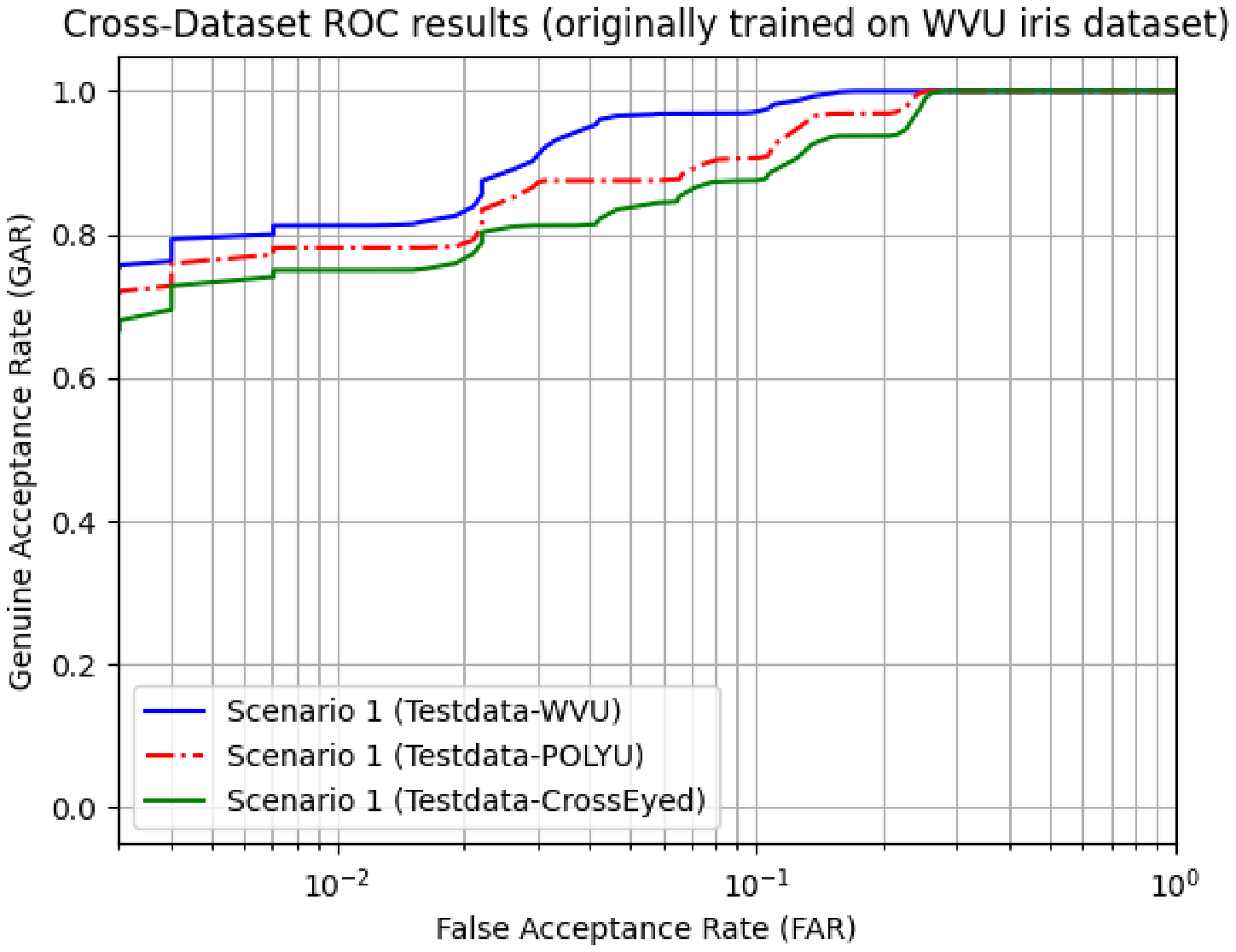}}
\subfloat[]{\includegraphics[width=7.5cm,height=5cm]{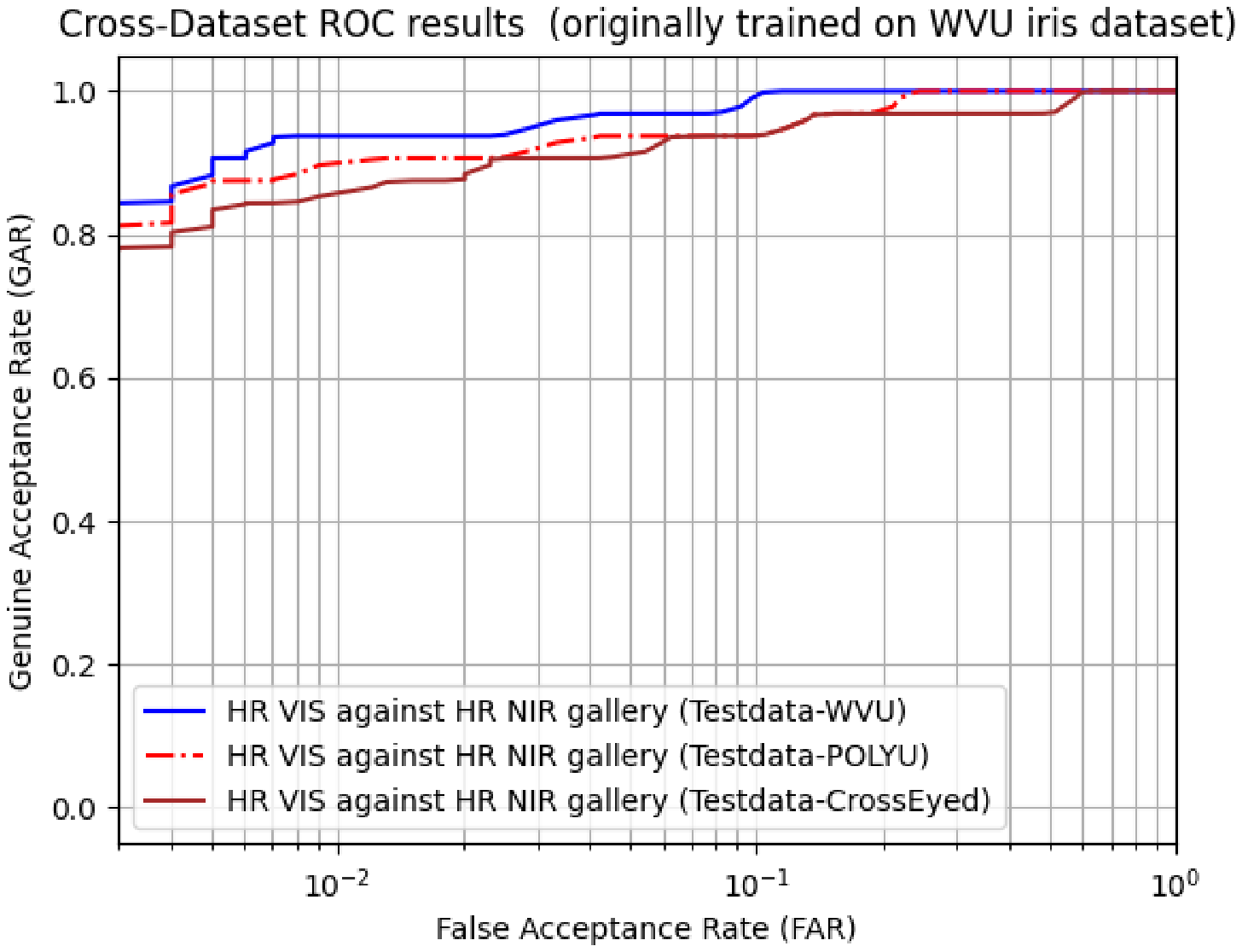}}
\caption{Comparative ROC results showing the cross-database matching of our approach (a) Scenario 1 with cGAN architecture (b) cpGAN architecture (matching the HR VIS iris probe against a HR NIR gallery) where both networks were trained only on the WVU face and iris dataset.}
\end{figure*}

\begin{figure*}%
\centering
\subfloat[]{\includegraphics[width=7.5cm,height=5cm]{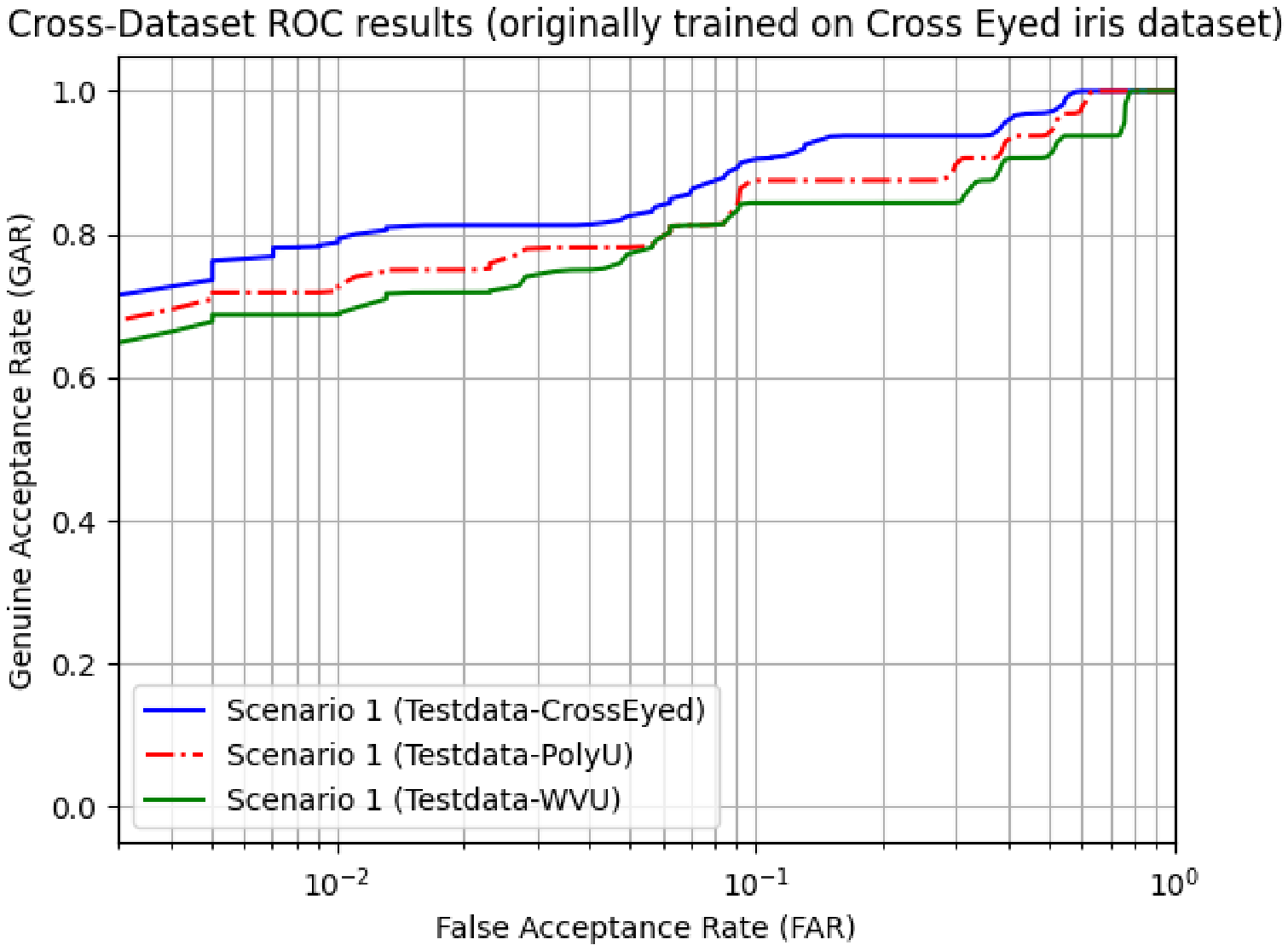}}
\subfloat[]{\includegraphics[width=7.5cm,height=5cm]{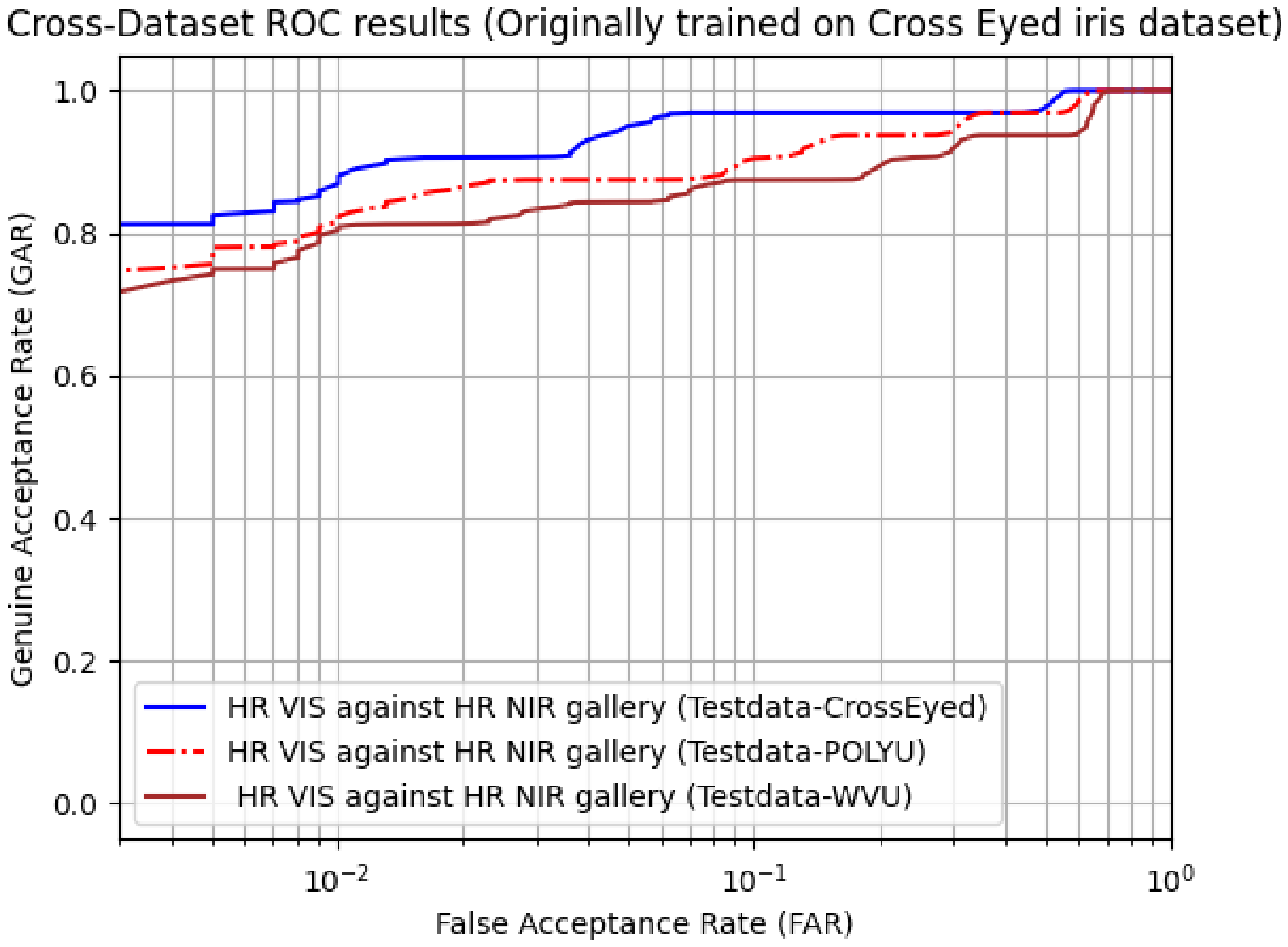}}
\caption{Comparative ROC results showing the cross-database matching of our approach (a) Scenario 1 with cGAN architecture (b) cpGAN architecture (matching the HR VIS iris probe against a HR NIR gallery) where both networks were trained only on the Cross-eyed-cross-spectral iris recognition dataset.}
\end{figure*}

\begin{table*}[t]
\centering
\caption{Cross-dataset matching performance evaluation. We trained both networks only on the PolyU bi-spectral dataset.}
\scalebox{0.90}{
\begin{tabular}{c|c|c|c|c|c}
\hline
Approach & Iris Comparison 
& Test Dataset &GAR@FAR=0.01&GAR@FAR=0.001&EER (\%)\\
\hline
\multirow{3}{6em}{\textbf{cpGAN(ours)}}
& \multirow{3}{6em}{HR VIS vs
HR NIR}& \textbf{PolyU Bi-Spectral}&\textbf{92.38}&\textbf{84.98}&\textbf{1.02}
\\
&& WVU Face and Iris&88.9&81.7&1.13
\\
&& Cross-eyed-cross-spectral&85.2&77.5&1.20
\\ \hline
\multirow{3}{6em}{\textbf{cGAN(ours)}}
& \multirow{3}{6em}{Scenario 1}& PolyU Bi-Spectral&80.5&70.1&1.5
\\
&& WVU Face and Iris&77.8&62.9&1.64
\\ 
&& Cross-eyed-cross-spectral&75.5&57.9&1.72
\\ \hline
\end{tabular}}
\end{table*}

\begin{table*}[t]
\centering

\caption{Cross-dataset matching performance evaluation. We trained both networks only on the WVU face and iris dataset.}

\scalebox{0.90}{
\begin{tabular}{ c|c|c|c|c|c}
\hline
Approach & Iris Comparison 
& Test Dataset &GAR@FAR=0.01&GAR@FAR=0.001&EER (\%)\\
\hline
\multirow{3}{6em}{\textbf{cpGAN(ours)}}
& \multirow{3}{6em}{HR VIS vs
HR NIR}&\textbf{WVU Face and Iris}&\textbf{93.0}&\textbf{84.0}&\textbf{0.90}
\\
&&PolyU Bi-Spectral&89.40&81.9&1.13
\\ 
&& Cross-eyed-cross-spectral&85.8&78.5&1.18
\\ \hline
\multirow{3}{6em}{\textbf{cGAN(ours)}}
& \multirow{3}{6em}{Scenario 1}&  WVU Face and Iris&80.8&75.1&1.54
\\
&&PolyU Bi-Spectral&79.6&68.9&1.60
\\
&& Cross-eyed-cross-spectral&76.0&67.5&1.66
\\ \hline
\end{tabular}}
\end{table*}

\begin{table*}[t]
\centering
\caption{Cross-dataset matching performance evaluation. We trained both networks only on the Cross-eyed-cross-spectral iris recognition dataset.}

\scalebox{0.90}{
\begin{tabular}{ c|c|c|c|c|c}
\hline
Approach & Iris Comparison 
& Test Dataset &GAR@FAR=0.01&GAR@FAR=0.001&EER (\%)\\
\hline
\multirow{3}{6em}{\textbf{cpGAN(ours)}}
& \multirow{3}{6em}{HR VIS vs
HR NIR}&  \textbf{Cross-eyed-cross-spectral}&\textbf{89.4}&\textbf{81.8}&\textbf{1.1}
\\
&& PolyU Bi-Spectral&82.30&74.80&1.21
\\
&& WVU Face and Iris&81.5&71.8&1.26
\\ \hline
\multirow{3}{6em}{\textbf{cGAN(ours)}}
& \multirow{3}{6em}{Scenario 1}& Cross-eyed-cross-spectral&80.09&70.1&1.54
\\
&& PolyU Bi-Spectral&71.5&68.9&1.75
\\
&& WVU Face and Iris&69.2&64.4&1.9\\
 \hline
\end{tabular}}
\end{table*}
\vspace{0.5cm}
\section{Cross-Database Performance Evaluation}
One of the most promising benefits of deep-learning-based iris recognition is its generalization capability, which offers high matching performance even when using the model trained on completely different iris database. Therefore, we also evaluate cross-database matching performance to validate the generalization capability of both of our approaches.

During this cross-database performance evaluation, first, we directly employ one of our models that has been trained on the PolyU bi-spectral database to ascertain the verification performance for the WVU face and iris database and Cross-eyed-cross-spectral iris recognition database without any fine-tuning. More specifically, we have used one dataset for training, and disjoint dataset for testing. Next, we follow the same technique to perform cross-database matching for the other two datasets:  we use a model trained on the WVU face and iris image database to evaluate the recognition performance for the PolyU and Cross-eyed database, and similarly, for a model that is trained using the Cross-eyed dataset. We maintain the same test-protocol as described for the respective databases in previous sections. For matching we consider only  \textcolor{blue}{Scenario 1} when evaluating the performance of the cGAN architecture. To report evaluation of the cpGAN network, we specifically consider the scenario where the HR VIS iris probe is matched against a HR NIR gallery. We have already introduced both of these scenarios in the earlier sections. 

The aim of this evaluation is to validate the generalization capability of our proposed frameworks when the target iris database has limited training samples. We show the comparative performance from the respective databases in \textcolor{blue}{Fig. 8-10} and report respective EER values in \textcolor{blue}{Table 5-7} from this cross-database performance evaluation. These results for the cross-database matching also indicate the performance improvement gained by employing our framework and reveal its generalization capability.

\section{Ablation Study}

\begin{table}[t]
\centering
\renewcommand{\arraystretch}{1}
\caption{Matching performance of our proposed cGAN using different hyperparameters settings on the PolyU Bi-Spectral test dataset.}
\scalebox{0.80}{\begin{tabular}{c|c}
\hline
Dataset & PolyU Bi-Spectral \\
\hline
Iris Comparison & Scenario 1 (HR VIS vs Synthesized VIS)\\
\hline
Hyperparameter Settings &GAR@FAR=0.01\hspace{0.35em} GAR@FAR=0.001\\
\hline
$\lambda_{1} = 10^{0}$, $\lambda_{2} = 2\times 10^{-3}$  &68.9\hspace{6em}54.6 \\
$\lambda_{1} = 10^{-2}$, $\lambda_{2} = 2\times 10^{-3}$  &75.7\hspace{6em}63.5\\
$\lambda_{1} = 10^{-4}$, $\lambda_{2} = 2\times 10^{-3}$  &78.1\hspace{6em}66.8\\
\scalebox{0.90}{\textcolor{black}{$\boldsymbol{\lambda_{1}
= 10^{-6}}$, $\boldsymbol{\lambda_{2} = 2\times 10^{-3}}$}}
&\textcolor{black}{\textbf{80.5}}\hspace{6em}\textcolor{black}{\textbf{70.1}} \\
$\lambda_{1} = 10^{-6}$, $\lambda_{2} = 2\times 10^{-2}$  &72.7\hspace{6em}61.4\\
$\lambda_{1} = 10^{-6}$, $\lambda_{2} = 2\times 10^{-1}$  &70.1\hspace{6em}58.5\\
\hline
\end{tabular}}
\end{table}

\begin{figure}%
\centering
{\includegraphics[width=7cm,height=5cm]{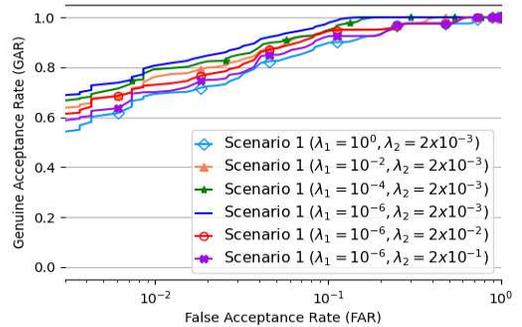}}
\caption{Comparative ROC results showing the sensitivity of matching performance on the hyperparameters of our proposed cGAN, when it is trained only on the PolyU Bi-Spectral dataset for Scenario 1.}
\vspace{-0.5cm}
\end{figure}

Training a GAN-based architecture is always difficult due to the GAN’s natural instability. Additional loss functions in guiding the GAN training can significantly improve the performance. However, these loss terms in the total combined loss are inconsistent on a numerical scale. Therefore, we use hyperparameters as weight factors to numerically balance the magnitude of different losses which accelerates the total loss convergence. To determine the optimal hyperparameters for our both cGAN and cpGAN models, we conduct an ablation study through changing the value of hyperparameters : $\lambda_{1}, \lambda_{2}$, and $\lambda_{3}, \lambda_{4}, \lambda_{5}$ adapted in equation (\ref{eq:19}) and (\ref{eq:20}), respectively. We have summarized the analysis in \textcolor{blue}{Tables 8-9}, and show the match performance in \textcolor{blue}{Figs. 11-12}.

\subsection{Hyperparameter Analysis}

\begin{table}[t]
\centering
\renewcommand{\arraystretch}{1}
\caption{Matching performance of our proposed cpGAN using different hyperparameters settings on the PolyU Bi-Spectral test dataset.}
\scalebox{0.78}{\begin{tabular}{c|c}
\hline
Dataset & PolyU Bi-Spectral \\
\hline
Iris Comparison & HR VIS vs HR NIR \\
\hline
Hyperparameter Settings &GAR@FAR=0.01\hspace{0.35em} GAR@FAR=0.001\\
\hline
$\lambda_{3} = 1$, $\lambda_{4} = 1$, $\lambda_{5} = 0.3$  &87.3\hspace{6em}74.2\\
$\lambda_{3} = 1$, $\lambda_{4} = 0.7$, $\lambda_{5} = 0.3$  &89.7\hspace{6em}78.9\\
$\lambda_{3} = 1$, $\lambda_{4} = 0.5$, $\lambda_{5} = 0.3$  &90.1\hspace{6em}81.8\\
\scalebox{0.90}{\textcolor{black}{$\boldsymbol{\lambda_{3}
= 1}$, $\boldsymbol{\lambda_{4} =0.3}$, $\boldsymbol{\lambda_{5} =0.3}$}}
&\textcolor{black}{\textbf{92.38}}\hspace{6em}\textcolor{black}{\textbf{84.98}} \\
$\lambda_{3} = 1$, $\lambda_{4} = 0.3$, $\lambda_{5} = 0.5$ 
&89.3\hspace{6em}76.6\\
$\lambda_{3} = 1$, $\lambda_{4} = 0.3$, $\lambda_{5} = 0.7$ 
&85.4\hspace{6em}74.0\\
$\lambda_{3} = 1$, $\lambda_{4} = 0.3$, $\lambda_{5} = 0.1$ 
&87.1\hspace{6em}71.9\\
\hline
\end{tabular}}
\end{table}

We evaluate the sensitivity of match performance when hyperparameters are varied across a range for training our proposed cGAN module. Training the cGAN with an $L_{2}$ term alone might lead to blurry results, since this loss penalizes the squared distance between ground truth outputs and synthesized outputs at pixel level. Since synthesized image quality is our top priority, we have added the ImageNet trained VGG-based perceptual loss, which is effective at generating realistic synthesized images by including more recognizable structure. 
Therefore, we keep the weight factor of the $L_{2}$ loss term 1 and train cGAN at $\lambda_{1}$ $\in$  $\{10^{0},10^{-6}\}$, and $\lambda_{2}$= $2\times 10^{-3}$, which are used as weight factors for adversarial loss term and perceptual loss term, respectively. We have also trained the network for a varied range of $\lambda_{2}$ $\in$ $\{2\times 10^{-3}$, $2\times 10^{-1}\}$, when $\lambda_{1}$ = $10^{-6}$. From the analysis of hyperparameters, as shown in \textcolor{blue}{Fig. 11}, and \textcolor{blue}{Table 8}, we notice that our proposed cGAN achieves the best matching performance for Scenario 1, when it is trained with $\lambda_{1}$ = $10^{-6}$, and $\lambda_{2}$ = $2\times 10^{-3}$ on the PolyU Bi-Spectral dataset. We have used this setting to perform all the experiments for cGAN and reported the obtained results in this paper. 

For training the cpGAN, we have considered additional constraints, such as $L_{2}$ loss and VGG-based perceptual loss along with adversarial, and contrastive loss functions. Since we have developed our 2nd  method to perform cross-spectral iris matching in the common embedded latent feature subspace, we put more emphasis on contrastive loss, which cares about the distance between genuine pairs and also penalizes mismatch between imposter pairs. Therefore, the weight factor for this loss term remains 1, and other hyperparameters have been changed to stabilize the cpGAN training, which allows it to converge faster, and thoroughly improve performance.

As seen in \textcolor{blue}{Fig. 12}, and \textcolor{blue}{Table 9}, we keep the adversarial weight factor, $\lambda_{3}$ = 1, when changing the values $\lambda_{4}$, and $\lambda_{5}$  from 0.3 to 1.0, which define weight factors for perceptual, and $L_{2}$ reconstruction loss term, respectively. From this ablation study, we have observed that $\lambda_{4}$= $\lambda_{5}$= 0.3 obtains the best matching accuracy, when the HR VIS iris probe is matched against the HR NIR iris gallery for the PolyU Bi-Spectral dataset (\textcolor{blue}{see Fig. 12 and Table 9}). For fair comparison, we have used these settings to train the cpGAN for other datasets and reported the results in this paper. 

\section{Limitation of the iris image acquisition method on the observed results}

The quality of iris images affects the matching performance of any iris recognition system, which indicates the significant role of the iris acquisition process.  It is the most initial part of any typical iris recognition system. During the acquisition of iris images, one must maintain an ISO standard iris image format (iris diameter has to be 150 pixels\textcolor{blue}{\cite{IRISREC}}), which is not easy to achieve in many data acquisition environments.  Most of the commercial iris image acquisition systems are designed to work at a close range and maintain a small operating distance, which is less than 1 meter\textcolor{blue}{\cite{liu2012review}}. Moreover, all of them need users’ cooperation. Therefore, it has become troublesome to capture iris images at a distance to generate low-resolution iris images in realistic environments. Therefore, there are no datasets available to study the effect cross-resolution and cross-spectral mismatch on  iris recognition systems in the literature. To overcome this limitation to some extent, the researchers developing state-of-the-art iris recognition systems have resized the original high-resolution iris images to their desired low-resolution images. In our work, we first apply a Gaussian filter and then resize the iris image using a bicubic interpolation method. We assume that these artificially-generated low-resolution images have similar characteristics as the original low-resolution images. However, we cannot certainly say that we would have achieved exactly similar performance if we used the original low-resolution images. We have tried to obtain low-resolution iris images as close as possible to a realistic setting. These results can be considered as a baseline for further improvement if the low-resolution iris images can be acquired in a realistic setting.

\begin{figure}%
\centering
{\includegraphics[width=7cm,height=5cm]{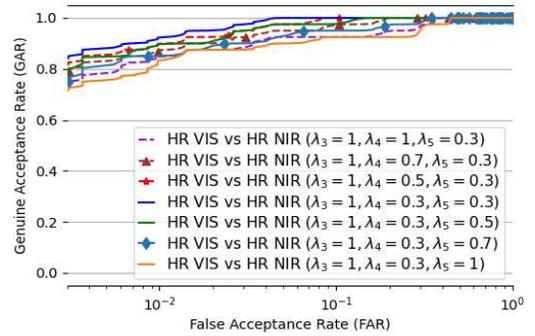}}
\caption{Comparative ROC results showing the sensitivity of matching performance of our proposed cpGAN, when it is trained only on the PolyU Bi-Spectral dataset for matching the HR VIS iris prob against a HR NIR gallery.}
\end{figure}


\section{Conclusion}
In this paper, we have described the development of two different deep learning frameworks for cross-spectral and cross- resolution iris recognition. While both  frameworks are developed based on domain transformation, one of them functions by translating from one domain to the another (NIR to VIS or vice versa), and the other framework transforms both domains to a latent embedding subspace. Briefly stated, in our first approach, we have introduced a domain translation network which can be considered as preprocessing step for any commercial off-the-shelf iris recognition system. In addition, we have proposed a new joint translation and super-resolution technique to address cross-resolution iris matching under the cross-domain problem. Experimental results on three publicly available cross-spectral datasets indicate the superiority of our proposed method over the earlier methods presented in the literature. This paper also investigates the domain invariant capability of our proposed cpGAN framework, which projects both the VIS and NIR iris texture features into a common latent embedding subspace to perform matching in the embedded domain. The goal of this network is to maximize the pair-wise correlation via contrastive loss during projection for more accurate cross-spectral iris matching. Results reported in Section 5 show significant improvement in the matching accuracy compared to other deep learning cross-spectral iris recognition algorithms. For instance, cpGAN achieves improvements of approximately 33\%, when compared to the results reported in \textcolor{blue}{\cite{wang2019toward}} for the PolyU Bi- Spectral dataset. Finally, we perform cross-database iris matching under the cross-spectral domain to evaluate the generalization capability of our methods.


%







{\small
\bibliographystyle{IEEEtran}
\bibliography{egbib}
}

\vspace{-0.65cm}
\begin{IEEEbiography}[{\includegraphics[width=1.0in,height=1.3in,clip]{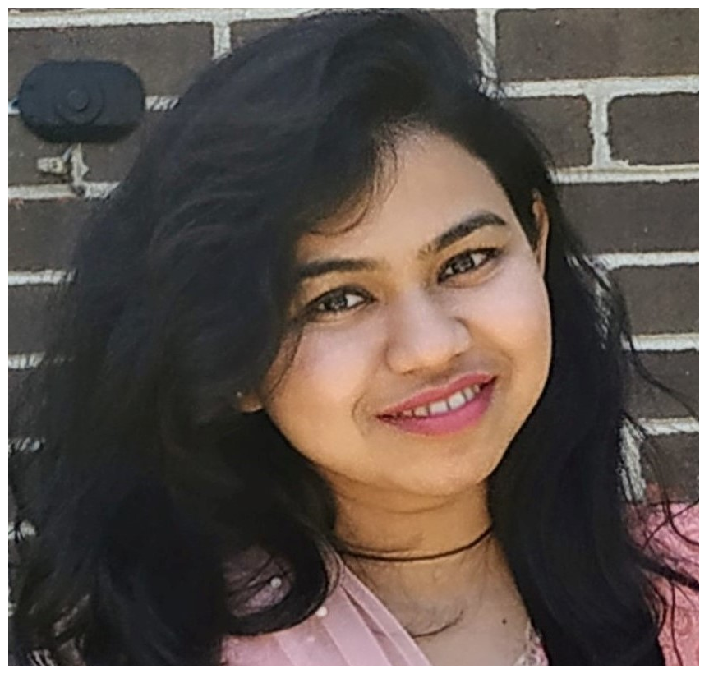}}]
{Moktari Mostofa} is currently a Ph.D. candidate at Lane Department of Computer Science and Electrical Engineering, West Virginia University, Morgantown,  WV,  USA. She received her B.Sc. degree in electrical and electronic engineering and M.Sc. degree in communication and signal processing from the University of Dhaka, Bangladesh in 2016 and 2018, respectively. Her research  includes mostly the applications of deep learning, machine learning, computer vision, biometrics, and image retrieval. 
\end{IEEEbiography}

\vspace{-1cm}
\begin{IEEEbiography}[{\includegraphics[width=1in,height=1.50in,clip,keepaspectratio]{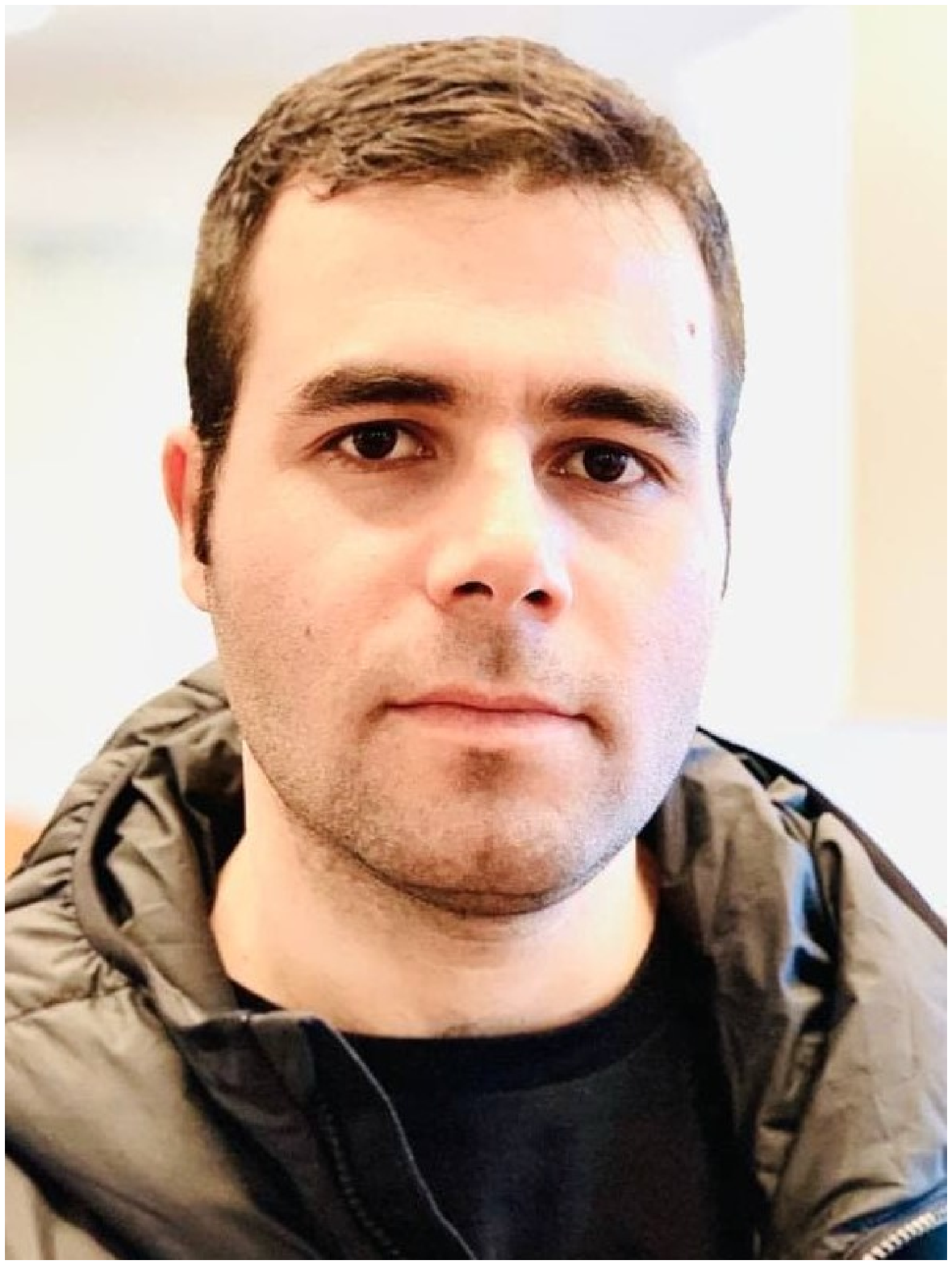}}]{Salman Mohamadi} is a Ph.D. candidate at West Virginia  University,  Morgantown,  WV, USA.  He received his B.Sc. degree and M.Sc. degree in electrical engineering from Shiraz University, Shiraz, Iran, and Amirkabir University of Technology (with honor), Tehran, Iran, respectively. His main focus is on development of machine learning and deep learning algorithms and their applications with Computer Vision, Biometrics and Bioinformatics.
\end{IEEEbiography}

\vspace{-1cm}
\begin{IEEEbiography}[{\includegraphics[width=1in,height=1.25in,clip]{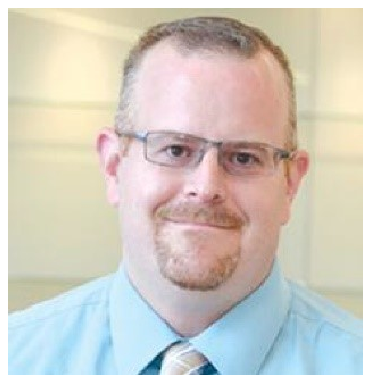}}]{Jeremy Dawson}, Associate Professor, joined the Lane Department of Computer Science and Electrical Engineering in the Fall of 2015. His background is in microelectronics and nanophotonics, and he has extensive experience in complex, multi-domain system integration and hardware system implementation. His current biometrics research efforts are focused on the creation of large-scale biometrics datasets that can be used to evaluate sensor operation and other human factors, as well as apply novel machine learning algorithms to solve human identification challenges.  Dr. Dawson also has extensive experience in micro and nanophotonic sensor platforms. His research in biosensors led to the identification of a need for new signal processing methodologies for DNA systems, which resulted in the first molecular biometrics (DNA) project funded through the WVU Center for Identification Technology Research (CITeR), an NSF IUCRC.
\end{IEEEbiography}

\vspace{-1cm}
\begin{IEEEbiography}[{\includegraphics[width=1.25in,height=1.25in,clip,keepaspectratio]{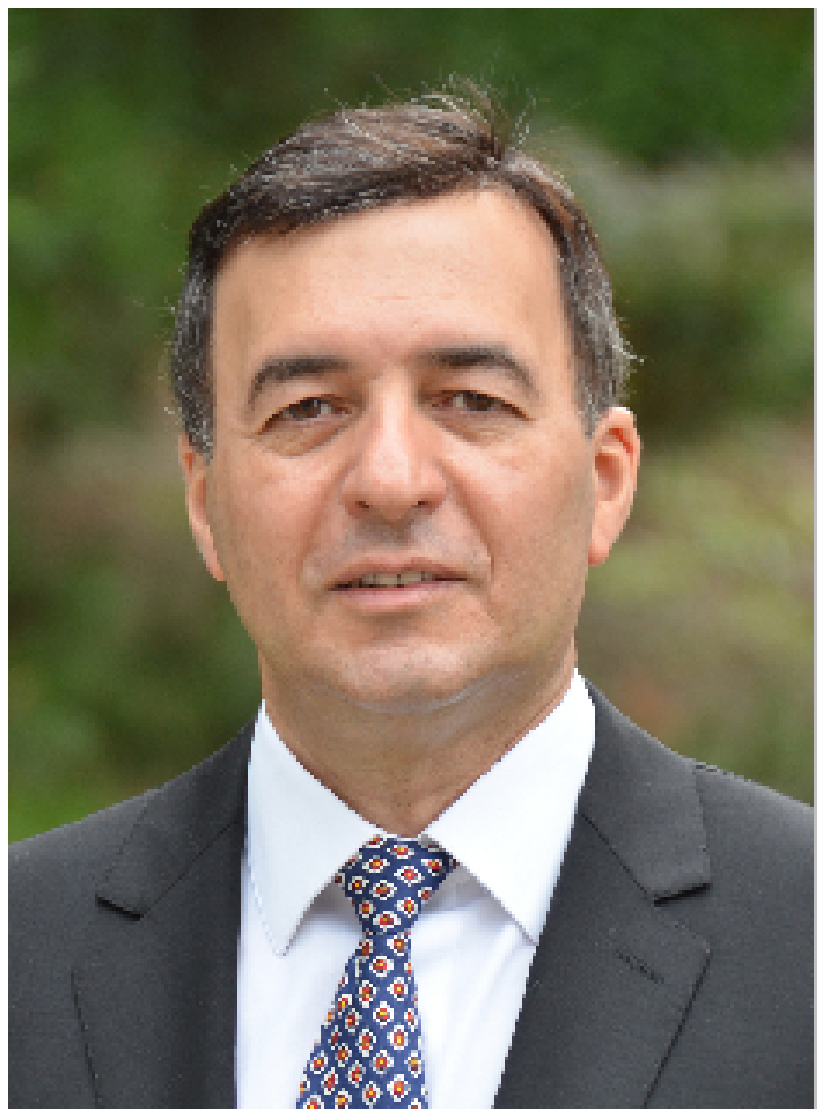}}]{Nasser M. Nasrabadi} (S’80 – M’84 – SM’92 – F’01) received the B.Sc. (Eng.) and Ph.D. degrees in electrical engineering from the Imperial College of Science and Technology, University of London, London, U.K., in 1980 and 1984, respectively. In 1984, he was with IBM, U.K., as a Senior Programmer. From 1985 to 1986, he was with the Philips Research Laboratory, New York, NY, USA, as a member of the Technical Staff. From 1986 to 1991, he was an Assistant Professor with the Department of Electrical Engineering, Worcester Polytechnic Institute, Worcester, MA, USA. From 1991 to 1996, he was an Associate Professor with the Department of Electrical and Computer Engineering, State University of New York at Buffalo, Buffalo, NY, USA. From 1996 to 2015, he was a Senior Research Scientist with the U.S. Army Research Laboratory. Since 2015, he has been a Professor with the Lane Department of Computer Science and Electrical Engineering. His current research interests are in image processing, computer vision, biometrics, statistical machine learning theory, sparsity, robotics, and neural networks applications to image processing. He is a fellow of ARL and SPIE and has served as an Associate Editor for the IEEE Transactions on Image Processing, the IEEE Transactions on Circuits, Systems and Video Technology, and the IEEE Transactions on Neural Networks.
\end{IEEEbiography}

\end{document}